# Learning Relaxation for Multigrid

**Dmitry Kuznichov**

# Learning Relaxation for Multigrid

Research Thesis

Submitted in partial fulfillment of the requirements
for the degree of Master of Science in Computer Science

**Dmitry Kuznichov**

**“There are only two mistakes one can make along the road to truth: not going all the way, and not starting.”** - *Buddha*

This research was carried out under the supervision of Prof. Irad Yavneh and Prof. Ron Kimmel, in the Faculty of Computer Science.

## Acknowledgements

First I would like to thank my supervisors, Prof. Irad Yavneh and Prof. Ron Kimmel. These two people opened the door for me to the world of academic research, and in their various instruction approaches gave me the opportunity to progress at my own pace and successfully complete my degree. In addition, I would like to thank all of the GIP team, who supported me before and throughout my degree studies.

I would also like to thank Dr. Dahn Katzen, Prof. Anatoly Golberg, and Dr. Yevgenia Apartsin (Yehudit Aperstein) of HIT, who were my first teachers in academia; my high school teacher Rachel Alon; and many others who have been my teachers over the years.

I would like to thank my friends Rémi Boissonnas, Inbal Keidar, Yonathan Galin-Galler, Yehonatan Knoll and Hila Kornis, who helped me with the linguistic editing of the thesis and made it much more readable to the supervisors and the rest of the world. Before my examination I presented my work to some other friends. I would like to thank them too, for their time, their attention and their feedback.

On a personal note, I would like to thank all the people who have accompanied me for many years, especially Moish. I thank my parents for their life-long love and endurance, which spurred my own endurance.

Thanks to all the challenging situations in my life!

The generous financial help of the Technion is gratefully acknowledged.

# Contents

# Abstract

During the last decade, Neural Networks (NNs) have proved to be extremely effective tools in many fields of engineering, including autonomous vehicles, medical diagnosis and search engines, and even in art creation. Indeed, NNs often decisively outperform traditional algorithms. One area that is only recently attracting significant interest is using NNs for designing numerical solvers, particularly for discretized partial differential equations. Several recent papers have considered employing NNs for developing multigrid methods, which are a leading computational tool for solving discretized partial differential equations and other sparse-matrix problems. We extend these new ideas, focusing on so-called relaxation operators (also called smoothers), which are an important component of the multigrid algorithm that has not yet received much attention in this context. We explore an approach for using NNs to learn relaxation parameters for an ensemble of diffusion operators with random coefficients, for Jacobi-type smoothers and for 4-Color Gauss-Seidel smoothers. The latter yield exceptionally efficient and easy to parallelize Successive Over Relaxation (SOR) smoothers. Moreover, this work demonstrates that learning relaxation parameters on relatively small grids using a two-grid method and Gelfand's formula as a loss function can be implemented easily. These methods efficiently produce nearly-optimal parameters, thereby significantly improving the convergence rate of multigrid algorithms on large grids.

# Chapter 1

# Introduction

In many areas of science and engineering, partial differential equations (PDEs) and systems are used to describe a broad range of phenomena, such as heat propagation, diffusion, electrodynamics, fluid dynamics, elasticity, quantum mechanics and much more. In all but a limited number of specific cases of partial differential equations that can be solved analytically, a numerical solution is the only practical option. In practice, the PDE is typically discretized on a (structured or unstructured) grid, resulting in a system of algebraic equations whose solution approximates the solution of the PDE at the grid nodes. For real-world problems, high accuracy is often required, and therefore the resulting system is very large and sparse [1]. For such problems, iterative methods are often the most effective. Notably, multigrid computational methods [BHM00, TOS01, BL11, Bra77, Yav06] are a leading iterative approach for solving problems of this type. Briefly described, multigrid methods employ a hierarchy of grids of different resolutions together with a simple classical iterative method. The iterative methods (called “relaxation operators” or “smoothers” in the Multigrid literature) include weighted (or damped) Jacobi and Gauss-Seidel for example. The goal of these operators is to efficiently smooth the error at the current iterate with respect to the grid. The smoothed error can then be approximated accurately on a coarser grid, and therefore eliminated inexpensively by the recursive multigrid process as described in Chapter 2.

Although multigrid methods were introduced several decades ago, designing such algorithms for new problems still requires expertise and experience. Often, there is no simple way to choose optimal multigrid components without trial and error. The idea of applying machine learning in general and NNs in particular for solving PDEs was proposed back in the early 1990s with works like [LK90], [PU92]. In [GGB$^+$19], the focus is on employing NNs to learn superior rules for prolongation operators, described below. In our work, we develop NNs in order to design the relaxation component of the multigrid algorithm with the aim of speeding up the solution process.

As in [GGB$^+$19], we focus on discretized diffusion equations in two dimensions (2D).

[1] In this report we focus on linear PDEs so the resulting algebraic systems are also linear.

A core element we explore is employing NNs to optimize the relaxation parameters with respect to an ensemble of random-coefficient diffusion operators. Our approach has three distinct advantages:

**Domain** Once the network finds optimal relaxation parameters, the values are efficient for an ensemble of diffusion coefficients, not just for one given PDE.

**Simplicity** The network uses unsupervised datasets to find optimal coefficients. As a result, transferring it to a new domain neither requires data annotation nor deep experience in partial differential equations. In addition the architecture of the chosen NN is simple and small, making the optimization process fast and cost effective.

**Scalability** Although the network is trained on sets of relatively small problems, we found that the optimal parameters for bigger problems are similar. Hence, values found for small problems provide good results also bigger problems.

The report is organized as follows. In the remainder of this chapter we report on related work. Then, in Chapter 2 we introduce the multigrid algorithm with specific focus on the relaxation operators we consider here, the model problem, and the method we use to learn optimal parameters. Chapter 4 describes the implementation and numerical experiments, and this is followed by a conclusions.

## 1.1 Related Work

### Machine learning and NN for solving Partial Differential Equations (PDE)

The idea of applying machine learning in general and NNs in particular for solving PDEs was proposed back in the early 1990s [LK90, PU92]. It gained popularity in the late 1990s with the work of [LLF98], which introduced a method for solving initial and boundary value problems using a NN. Not long thereafter, [LLP00] introduced radial basis function networks that yielded accurate results for PDEs on domains with irregular boundaries. But it was nearly 20 years later that the field attracted widespread interest and a surge of publications appeared.

Physics-informed neural networks (PINN) is the name given to a technique proposed in [RPK19], which has become widely accepted for the type of PDE solvers investigated by the authors. The first part of the paper is dedicated to data-driven solutions. It focuses on computing the solutions of general forms of PDEs with supervised training of a deep NN feeding into the component which implements the physical equations (the "physics-informed part"). The supervised training is conducted with a loss function that encompasses the losses of data initialization, boundary conditions and the PDE at a finite set of collocation points. A second part is dedicated to data-driven discovery.

It consists of finding parameters of a particular PDE given a sample solution. The outcome of a deep NN is also fed into a physics-informed part, but unlike the data-driven solution approach, some of the parameters in this physics-informed part are trainable. The paper illustrates the approaches by giving examples based on both continuous and discrete time models.

Another well-known paper, [WY18], introduces *Deep Ritz methods* (DRM). It implements a similar principle of data-driven solution but with a different loss function, using the Ritz method rather than the collocation method. Related work was presented in [KZK21] and [SS18] using other loss functions, respectively, the Galerkin method for hp-Variational Physics-Informed Neural Networks (hp-VPINNs) and the Petrov-Galerkin method for the Deep Galerkin Method (DGM). Having gained recognition, PINN spawned further research, e.g, [BS19, SAR20, ZBYZ20, MFK21].

Another significant recent approach is *Convolutional Neural Networks* (CNN) used as PDE solvers, mostly employing encoder-decoder architectures. In such architectures, the data from the input is forced through a bottleneck that significantly reduces its dimensions (encoder) and is then expanded again (decoder). Only the most salient features of the input are kept in the process and exposed in the output. These methods have become influential in many fields, such as data analysis, computer vision and natural language processing, where they are typically used for dimensional reduction, information retrieval, anomaly detection, denoising and more. Deep encoder-decoder architectures make it possible to train NNs to compute approximate solutions much faster without significantly increasing errors. For example, [GLI16, KLY20, ZZ18, BAP+19] take this approach. In [GLI16], convolutional neural networks are shown to be a faster alternative than classic computational fluid dynamics simulations. These CNNs can estimate the velocity field up to two orders of magnitude faster. The aim of [KLY20] is similar to that of the second part of the original PINN paper [RPK19]. It is likewise a data-driven solution which takes in a sample solution and finds parameters of a known PDE. But unlike PINN, this approach uses a decoder architecture without prior knowledge of the physical equation. The research described in [ZZ18] aims to improve uncertainty quantification and propagation of problems represented by stochastic PDEs with the help of surrogate models. The performance of the proposed models is assessed on flows in heterogeneous media with permeability as an input (encoder) and flow and pressure fields as outputs (decoder). Similarly to [ZZ18], [BAP+19] predicts velocity and pressure fields in unseen flow conditions using CNN. It focuses on flow solutions over airfoil shapes of Reynolds-averaged Navier-Stokes equations. These solvers are used in uncertainty quantification, computational fluid dynamics and other applications.

Other approaches have been proposed recently. [HJW18] presents a practical algorithm for solving nonlinear partial differential equations in very high dimensions, and [BN18] uses deep feed-forward NNs to approximate solutions of partial differential equations in complex geometries. In [SYS03], a new signal processing method is utilized to solve PDEs numerically using a NN. In this paper, the theoretical aspects of this approach are investigated, and it is shown that the numerical computation can be formulated as a machine learning problem and implemented by a supervised NN. [Mis18] proposes a machine learning framework to accelerate numerical computations of time-dependent ODEs and PDEs. The method is based on recasting existing numerical methods as NNs, with a set of trainable parameters. [BRFF18] introduces a new way to optimize complex shapes quickly and accurately. The key to making this approach practical is remeshing the original shape using a polycube map, which makes it possible to perform the computations on GPUs instead of CPUs. In [WJL18], nonlinear ODEs and PDEs are solved by a universal rule-based self-learning approach using deep reinforcement learning. [TSD+17] investigates the feasibility of applying deep learning techniques to solve Poisson's equation.

### NNs for the development of Multigrid solvers

The vast majority of multigrid developments is performed with traditional tools. Recent examples include [BHM+20], which uses local Fourier analysis for robust optimization, whereas [TMGV20] and [FKMW21] focus on improving relaxation.

Nevertheless, the number of papers using machine learning for developing multigrid methods is steadily growing. The first paper to consider this is [KDO17], which proposes a method to optimize restriction and prolongation operators for geometric multigrid solvers. This approach is restricted to constant-coefficient problems on simple geometries, and the method needs to be retrained for each new problem. A far more ambitious approach is presented in [GGB+19], where a NN is trained once to learn rules for creating prolongation operators. The network can then be used for an entire ensemble of diffusion problems with random coefficients on regular grids, and is shown to yield faster convergence than the state-of-the-art Black Box multigrid method. This work is extended in [LGM+20] to design prolongation operators for Algebraic Multigrid (AMG) that can handle unstructured problems which are not necessarily discretizations of PDEs (for example, graph Laplacians or Markov chains). In [MHLR21], a deep neural network multigrid solver (DNN-MG) is introduced. This solver uses multigrid to provide solutions on coarse levels while the NN improves this solution on fine grids. This approach only requires about half the computation time to solve 2D laminar flows around an obstacle (Navier-Stokes equations) compared to standard multigrid methods.

CNN-based optimization of multigrid smoothers is presented in [HLX21]. A set of smoothers results from the supervised training of finite sequences of convolution kernels. Each smoother optimizes the convergence rate for a given grid level. The NNs are trained on specific PDEs (anisotropic rotated Laplacian problems and variable coefficient diffusion problems) with a supervised loss function. One of the benefits of these smoothers is that, after they have been trained on small-scale problems, they can be applied to large-scale problems as well.

# Chapter 2

# Multigrid Relaxation with NN

A common way to (approximately) solve linear partial differential equations is to discretize them using a Finite-Difference (FD) or Finite-Element Method (FEM), obtaining a square linear system of the form

$$Au = f, \tag{2.1}$$

where $A$ is a square matrix of size $n \times n$, and $u$ and $f$ are vectors of size $n$, representing the unknown solution vector and the given forcing vector, respectively. The discretization step will be addressed in more detail in section 2.2. Here we focus on the iterative solution of the linear system, assuming $A$ is nonsingular.

## 2.1 Iterative Methods

Solving large linear systems like (2.1) is, in the general case, a computationally expensive task, requiring $O(n^3)$ operations. When the matrix $A$ is sparse, however, as in the case of discretized partial differential equations, iterative methods are attractive. Classical simple iterative schemes, such as Gauss-Seidel or Jacobi relaxations, are generally of the form

$$u^{(k+1)} = u^{(k)} + M(f - Au^{(k)}), \tag{2.2}$$

or, with $S = I - MA$, where $I$ is the identity matrix,

$$u^{(k+1)} = Mf - Su^{(k)}, \tag{2.3}$$

where $u^{(k)}$ denotes the approximation to $u$ after $k$ iterations starting with an initial approximation $u^{(0)}$, and $M$ is some approximate inverse of $A$ that is easy to compute[1].

Denote by

[1]More precisely, the system $M^{-1}u = g$ must be easy to solve for any vector $g$ of size $n$.

$$e^{(k)} = u - u^{(k)}, \tag{2.4}$$

the error in our approximation to $u$ after the $k$th iteration. Subtracting both sides of (2.3) from $u$ and substituting $Au$ for $f$, we obtain the relation

$$e^{(k+1)} = Se^{(k)}. \tag{2.5}$$

A sufficient and necessary condition for convergence of the iterative method to the solution $u$ for any initial guess $u^{(0)}$ is given by

$$\rho(S) < 1, \tag{2.6}$$

where $\rho(S)$ is the spectral radius of matrix $S$. Another important metric of current error is the residual $r^{(k)}$ after the $k$th iteration, which is given by

$$r^{(k)} = f - Au^{(k)} = Ae^{(k)}. \tag{2.7}$$

From (2.7), it can be seen that there is a direct connection between the convergence of $e$ and of $r$ to zero.

It is common to introduce a scalar *relaxation parameter* $\omega$ into (2.2) as follows:

$$u^{(k+1)} = u^{(k)} + \omega M(f - Au^{(k)}), \tag{2.8}$$

where, for $\omega = 1$, (2.8) is reduced to (2.2). It is of course possible to integrate $\omega$ into matrix $M$, but $\omega$ is usually written explicitly, and then $S = I - \omega MA$. If (2.8) is used as a stand-alone solver, the objective then would be to select $\omega$ which minimizes $\rho(S)$. However, the optimal $\omega$ depends on certain eigenvalues of $MA$, which are not known except in some special cases.

It is common practice to decompose the matrix $A$ into the sum of its diagonal and lower and upper triangular components. Denoting by $D$ the diagonal matrix with values given by the diagonal of $A$, and by $L$ ($U$) the lower (upper) triangular matrix comprised of the corresponding elements of $A$, we have $A = D + L + U$. We also denote $L_+ = L + D = A - U$.

Two well-known iterative methods are Jacobi and Gauss-Seidel. In the Jacobi method, $M$ in Eq. (2.2) is equal to $D^{-1}$, and the iteration matrix is $S = I - D^{-1}A = -D^{-1}(L + U)$. In the Gauss-Seidel method, $M = L_+^{-1}$, so $S = I - L_+^{-1}A = -L_+^{-1}U$. Written out explicitly, these relations take the following form. Let

$$A = \begin{bmatrix} a_{11} & a_{12} & \cdots & a_{1n} \\ a_{21} & a_{22} & \cdots & a_{2n} \\ \vdots & \vdots & \ddots & \vdots \\ a_{n1} & a_{n2} & \cdots & a_{nn} \end{bmatrix}, u^{(k)} = \begin{bmatrix} u_1^{(k)} \\ u_2^{(k)} \\ \vdots \\ u_n^{(k)} \end{bmatrix}, f = \begin{bmatrix} f_1 \\ f_2 \\ \vdots \\ f_n \end{bmatrix}. \tag{2.9}$$

Then the $i$th element of the approximation to $u$ after iteration $k+1$ is $u_i^{(k+1)}$. For the Jacobi method it is calculated as

$$u_i^{(k+1)} = \frac{1}{a_{ii}} \left( f_i - \sum_{j \neq i} a_{ij} u_j^{(k)} \right) , \tag{2.10}$$

whereas for Gauss-Seidel it reads

$$u_i^{(k+1)} = \frac{1}{a_{ii}} \left( f_i - \sum_{j=1}^{i-1} a_{ij} u_j^{(k+1)} - \sum_{j=i+1}^{n} a_{ij} u_j^{(k)} \right) . \tag{2.11}$$

The advantage of Gauss-Seidel is that it uses updated values of $u$ on the fly, which usually speeds up convergence compared to Jacobi, where the values are updated only at the end of each iteration. The effect of this may be significant [Maz16]. The computational advantage of the Jacobi method, on the other hand, is that the computations in each iteration can easily be done in parallel, while for Gauss–Seidel this is not generally true. Conditions on matrix $A$ for the convergence of Jacobi and Gauss-Seidel iterative methods are given in Appendix A.1

The weighted versions of Jacobi ($M = \omega D^{-1}$) and Gauss-Seidel methods are derived from (2.8). We remark that usually $0 < \omega < 1$ is required for Jacobi, and it is then often called weighted Jacobi, whereas $\omega > 1$ is commonly best for Gauss-Seidel, and it is often called Successive Over-Relaxation (SOR). The latter can be written explicitly as follows.

$$u_i^{(k+1)} = (1-\omega) u_i^{(k)} + \omega \cdot \underbrace{\frac{1}{a_{ii}} \left( f_i - \sum_{j=1}^{i-1} a_{ij} u_j^{(k+1)} - \sum_{j=i+1}^{n} a_{ij} u_j^{(k)} \right)}_{Gauss-Seidel} . \tag{2.12}$$

Because weighted Jacobi and SOR are generalizations of Jacobi and Gauss-Seidel, respectively, their convergence rates are at least as good as these, given appropriate relaxation parameters. By way of "sanity check", when $\omega = 1$, the SOR method is exactly equal to Gauss-Seidel as mentioned earlier, and when $\omega = 0$, the method just keeps the current value of the approximate solution $u^{(k)}$ unchanged. Eq. (2.12) shows that $\omega$ is in essence an arbitrator between the accuracy of the approximation at the $k$th step and the new value given by the standard Gauss-Seidel method of Eq. (2.11). $\omega$ values that are closer to 0 indicate a "distrust" in the correction. Values of $\omega$ close to but smaller than 1 (under-relaxation) are typically chosen when a more conservative iteration is required, whereas values of $\omega$ above 1 (over-relaxation) indicate a high degree of trust placed in the direction of change given by the Gauss-Seidel iteration. The use of iterative methods as so-called smoothers in multigrid solvers is detailed in section 2.3.

An alternative iterative method, which is far less dependent on careful tuning of a relaxation parameter, is the Sparse Approximate Inverse (SPAI) method [BG02], whereby $M$ is optimized subject to sparsity constraints. An ideal choice of $M$ would be one that minimizes $\rho(S)$ given some prescribed (or adaptively chosen) sparsity patterns of $M$. This is a hard problem to solve in general. In SPAI, [BG02] replaces this by the minimization of the Frobenius norm (which bounds the spectral radius from above).

$$M = \text{argmin}_{\tilde{M}} ||I - \tilde{M}A||_F, \tag{2.13}$$

subject to some sparsity constraints on $M$.

## 2.2 The Model Problem

Our model problem in this research is the classical elliptic diffusion partial differential equation,

$$-\nabla \cdot (\mathbf{G}(\mathbf{r})\nabla \mathbf{U}(\mathbf{r})) = \mathbf{F}(\mathbf{r}), \tag{2.14}$$

where $\mathbf{G}$ denotes the positive diffusivity, $\mathbf{F}$ is a given forcing function, $\mathbf{U}$ is the solution we are seeking, and $\mathbf{r}$ is the continuous space variable. In case $\mathbf{G} = 1$, (2.14) is reduced to the Poisson equation. In this report we restrict our discussion to two spatial dimensions, denoted 2D. The continuous diffusion equation (2.14) is then discretized on a grid comprised of $m \times m$ cells with doubly periodic "boundary conditions" (Fig. 2.2b). The mesh-sizes of the rectangular cells are denoted by $h_x$ and $h_y$. The case $h_x = h_y = h$ is referred to as isotropic, whereas it is called anisotropic for $h_x \neq h_y$. We write the resulting discrete diffusion equation as

$$-\nabla^h \cdot (\mathbf{g}\nabla^h \mathbf{u}) = \mathbf{f}, \tag{2.15}$$

where $\mathbf{g}$, $\mathbf{u}$ and $\mathbf{f}$ are 2D arrays whose elements are the discrete approximations of $\mathbf{G}$, $\mathbf{U}$ and $\mathbf{F}$ at the nodes of the grid, and the $h$ superscripts indicate that these are discrete approximations to the divergence and gradient operators. This is a convenient way to represent the sparse linear system of the form (2.1) that results from the discretization. Adopting the conventions of Fig. 2.1, we discretize the diffusion operator using bilinear

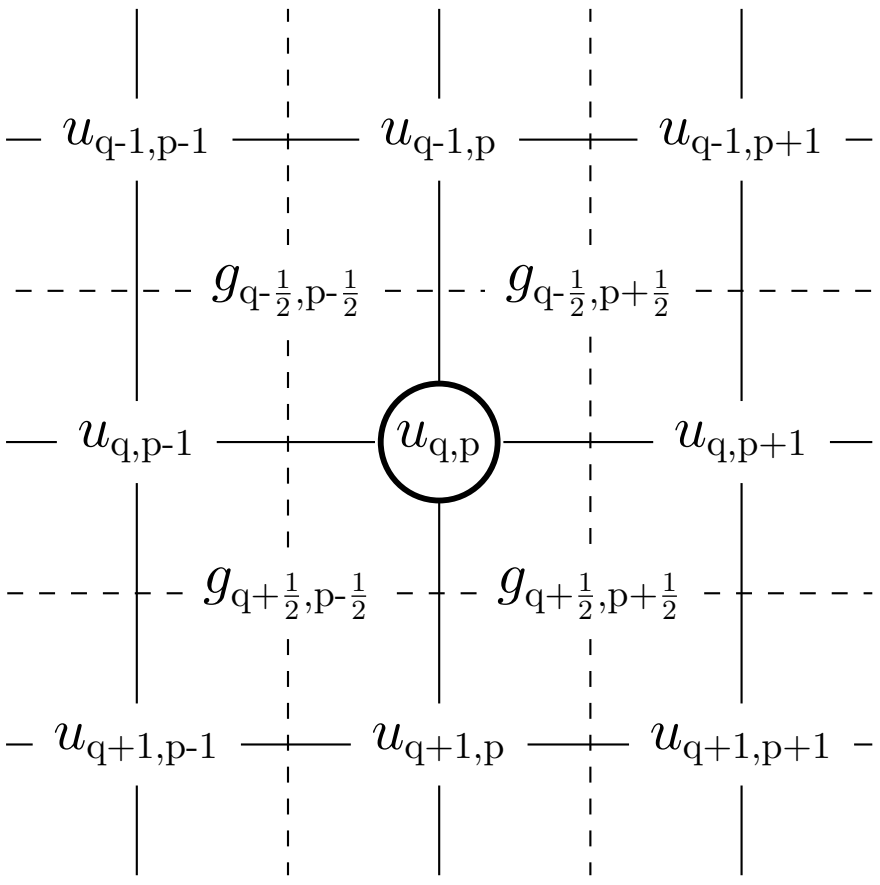

Figure 2.1: A small portion of the grid in the isotropic case ($h_x = h_y$), demonstrating the location of the unknown $u$ variables at the nodes and the given $g$ values at cell centers.

finite elements, obtaining at grid-node $(q, p)$,

$$\begin{aligned}
&-\frac{1}{6}\left(\frac{1}{h_y^2}+\frac{1}{h_x^2}\right)\left(g_{q-\frac{1}{2},p-\frac{1}{2}}u_{q-1,p-1}+g_{q-\frac{1}{2},p+\frac{1}{2}}u_{q-1,p+1}+g_{q+\frac{1}{2},p+\frac{1}{2}}u_{q+1,p+1}+g_{q+\frac{1}{2},p-\frac{1}{2}}u_{q+1,p-1}\right)\\
&+\frac{1}{6}\left(\frac{1}{h_x^2}-\frac{2}{h_y^2}\right)\left(\left(g_{q-\frac{1}{2},p-\frac{1}{2}}+g_{q-\frac{1}{2},p+\frac{1}{2}}\right)u_{q-1,p}+\left(g_{q+\frac{1}{2},p+\frac{1}{2}}+g_{q+\frac{1}{2},p-\frac{1}{2}}\right)u_{q+1,p}\right)\\
&+\frac{1}{6}\left(\frac{1}{h_y^2}-\frac{2}{h_x^2}\right)\left(\left(g_{q-\frac{1}{2},p+\frac{1}{2}}+g_{q+\frac{1}{2},p+\frac{1}{2}}\right)u_{q,p+1}+\left(g_{q+\frac{1}{2},p-\frac{1}{2}}+g_{q-\frac{1}{2},p-\frac{1}{2}}\right)u_{q,p-1}\right)\\
&+\frac{1}{3}\left(\frac{1}{h_y^2}+\frac{1}{h_x^2}\right)\left(g_{q-\frac{1}{2},p-\frac{1}{2}}+g_{q-\frac{1}{2},p+\frac{1}{2}}+g_{q+\frac{1}{2},p+\frac{1}{2}}+g_{q+\frac{1}{2},p-\frac{1}{2}}\right)u_{q,p}=f_{q,p}.
\end{aligned} \tag{2.16}$$

in the isotropic case, we have

$$\begin{aligned}
&\underbrace{\frac{-1}{3h^2}g_{q-\frac{1}{2},p-\frac{1}{2}}}_{=a_{x,x-m-1}}u_{q-1,p-1}+\underbrace{\frac{-1}{3h^2}g_{q-\frac{1}{2},p+\frac{1}{2}}}_{=a_{x,x-m+1}}u_{q-1,p+1}+\underbrace{\frac{-1}{3h^2}g_{q+\frac{1}{2},p+\frac{1}{2}}}_{=a_{x,x+m+1}}u_{q+1,p+1}+\underbrace{\frac{-1}{3h^2}g_{q+\frac{1}{2},p-\frac{1}{2}}}_{=a_{x,x+m-1}}u_{q+1,p-1}\\
&+\underbrace{\frac{-1}{6h^2}\left(g_{q-\frac{1}{2},p-\frac{1}{2}}+g_{q-\frac{1}{2},p+\frac{1}{2}}\right)}_{=a_{x,x-m}}u_{q-1,p}+\underbrace{\frac{-1}{6h^2}\left(g_{q+\frac{1}{2},p+\frac{1}{2}}+g_{q+\frac{1}{2},p-\frac{1}{2}}\right)}_{=a_{x,x+m}}u_{q+1,p}\\
&+\underbrace{\frac{-1}{6h^2}\left(g_{q-\frac{1}{2},p+\frac{1}{2}}+g_{q+\frac{1}{2},p+\frac{1}{2}}\right)}_{=a_{x,x+1}}u_{q,p+1}+\underbrace{\frac{-1}{6h^2}\left(g_{q+\frac{1}{2},p-\frac{1}{2}}+g_{q-\frac{1}{2},p-\frac{1}{2}}\right)}_{=a_{x,x-1}}u_{q,p-1}\\
&+\underbrace{\frac{2}{3h^2}\left(g_{q-\frac{1}{2},p-\frac{1}{2}}+g_{q-\frac{1}{2},p+\frac{1}{2}}+g_{q+\frac{1}{2},p+\frac{1}{2}}+g_{q+\frac{1}{2},p-\frac{1}{2}}\right)}_{=a_{x,x}}u_{q,p}=f_{q,p}.
\end{aligned} \tag{2.17}$$

Eq. (2.16) represents the diffusion equation at node $q, p$ of the 2D grid, with $f_{p,q}$ denoting a local average of $\mathbf{f}$ values with weights given by the finite element discretization. The solution $\mathbf{u}$ satisfies Eq. (2.15) if and only if it satisfies the entire set of

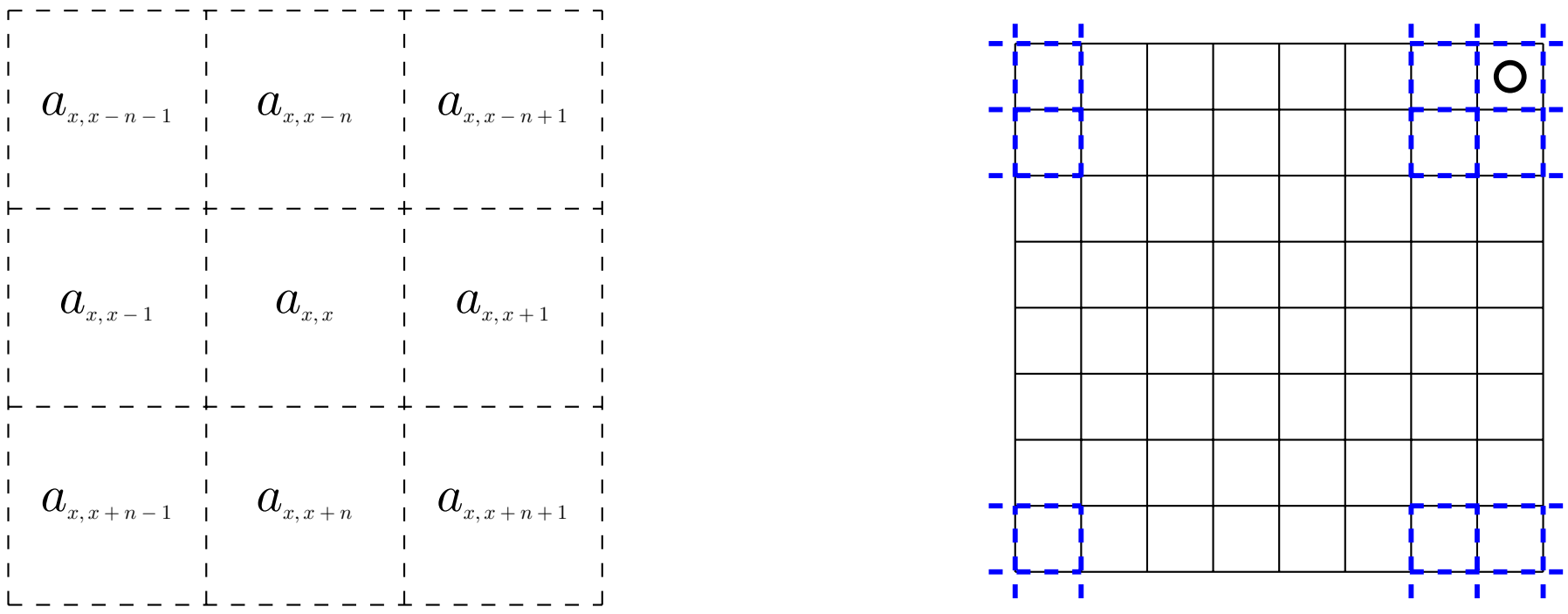

(a) A stencil which links the discretized 2D space $\mathbf{r}$ (grid) and the indices of the non-zero values in a given row $x$ of matrix $A$.

(b) Illustration of the doubly periodic boundary conditions on the discretized 2D space $\mathbf{r}$.

Figure 2.2

equations for all nodes of the 2D grid (i.e. $\forall q, p \in \{0, ..., m-1\}$). As is well known, the most common way to solve a sparse set of linear equations such as this is by numerical linear algebra tools, for example any of the iterative methods presented in section 2.1. The required steps for transferring (2.17) to the system of linear equations of the form (2.1) include: replacing $\mathbf{u}$, $\mathbf{f}$ ($m \times m$ on 2D grid) with $u$, $f$ vectors (of size $n = m^2$) and finding the appropriate indices for the coefficients ($a_{x,y}$) of matrix $A$ (of size $m^2 \times m^2$). The first step is nothing but a flattening of $\mathbf{u}$ and $\mathbf{f}$ in row-major order. The $x$ indices of the coefficients of $A$ are given by $x = q \cdot m + p$ and the principle of calculation of the $y$ indexes is presented in Figure 2.2a. This step requires special attention for the nodes located around the borders as illustrated in Figure 2.2b. The resulting $A$ is a symmetric positive semi-definite (SPSD) matrix, which is in general large and sparse and contains nine nonzero values per row. Note that the values of $u_{i,j}$ and $f_{i,j}$ with two indices (ref. equation (2.17)) correspond to node $i, j$ on the 2D grid while $u_i$ and $f_i$ with one index belong to the vectors of the linear system of Eq. (2.1) and satisfy the following equation (with appropriate modifications at the borders):

$$\begin{aligned} &a_{x,x-n-1}u_{x-n-1} + a_{x,x-n+1}u_{x-n+1} + a_{x,x+n+1}u_{x+n+1} + a_{x,x+n-1}u_{x+n-1} \\ &+ a_{x,x-n}u_{x-n} + a_{x,x+n}u_{x+n} + a_{x,x+1}u_{x+1} + a_{x,x-1}u_{x-1} + a_{x,x}u_x = f_x. \end{aligned} \tag{2.18}$$

For example, the value of $u$ at row $x$ when the Jacobi iterative method is used (2.11), is given by

$$u_x^{(k+1)} = \frac{1}{a_{x,x}} \left( f_x - \sum_{j \in l} a_{x,j} u_j^{(k)} \right), \tag{2.19}$$

where $l$ is the set of indices of all the nonzeros in row $x$ of $A$.

## 2.3 Multigrid

Linear systems that result from the discretization of partial differential equations, as we consider here, are typically large, very sparse, and highly ill-conditioned (see subsection 2.2). For such systems, simple iterative methods described in subsection 2.1 are cheap to apply because the number of operations per iteration is bounded by the number of variables times a small constant. However, they tend to converge very slowly. To greatly accelerate convergence we use multigrid methods which employ a hierarchy of grids, from the original fine computational grid down to a very coarse grid. The components of multigrid methods are relaxation (also called smoothing), restriction and prolongation. These can be adjusted to improve the overall performance of the method. In many cases, common simple iterative methods as described above are very efficient at smoothing the error relative to the grid, thus making it possible to approximate it accurately on a coarser grid, and to solve a suitable linear system for this coarse-grid approximation. The coarse-grid solution is then interpolated back to the fine grid using a prolongation operator $P$, and used for correcting the fine-grid error. The matrix $P$ may also be used to project the fine-grid system to the coarse grid, discussed below. In this work we restrict $A$ to be symmetric, and correspondingly use $P^T$ for the restriction. The prolongation operators ($P$) which we use are standard out-of-the-box schemes.

The process of "coarsening" the problem by approximating it on a coarser grid is performed recursively, traversing the hierarchy of grids from finest to coarsest and back again. One such back and forth traversal constitutes a single multigrid iteration, called a cycle. These cycles are repeated until some appropriate convergence criterion is satisfied.

The multigrid cycle exploits an entire hierarchy of grids whereas the two-grid version of the cycle employs only the original finest grid and a single coarse grid on which the coarse-grid problem is solved exactly. This version is generally used only for study purposes, as it is typically inefficient as a solver because the coarse-grid direct solution is still expensive. And yet, it is useful for analysis and, more importantly in the framework of this work, for learning optimal multigrid parameters. We next describe the two-grid cycle, which is the core building-block of the multigrid cycle.

Each iteration of the two-grid algorithm starts from the fine grid and executes a few repetitions of the prescribed relaxation as explained in subsection 2.1 (typically just one or two iterations). The error that remains at this stage is (assumed to be) smooth relative to the grid and can thus be approximated well on the coarse grid. The coarse grid operator is the so-called Galerkin operator which is applied to matrix $A$ to give $A_c$, the coarse grid version of matrix $A$,

$$A_c = P^T A P. \tag{2.20}$$

The right-hand side of the coarse-grid system at the $k$th iteration is the restricted residual, $P^T r^{(k)}$, with $r^{(k)} = f - Au^{(k)} = Ae^{(k)}$, as defined in Eq. (2.7), the residual remaining after the relaxation. The coarse grid problem,

$$A_c e_c^{(k)} = P^T A P e_c^{(k)} = P^T r^{(k)} = r_c, \tag{2.21}$$

is then solved, and the resulting coarse-grid correction is prolongated and added to the current fine-grid approximation, $u^{(k)} \leftarrow u^{(k)} + Pe_c^{(k)}$. Finally, more relaxations are executed (again, typically just one or two). This completes the two-grid cycle.

**Algorithm 2.1** Two-Grid Cycle

**Input:**
- i Discretization matrix $A$
- ii Initial approximation $u^{(0)}$
- iii Right-hand side $f$
- iv $P$ matrix for Prolongation & Restriction
- v $M$ matrix for Relaxation
- vi $\nu_1$ and $\nu_2$ number of relaxation iterations before and after coarse grid correction
- vii Counter $k = 0$
- viii Residual tolerance $\delta$

**Output:**
- i Solution $u^{(k)}$

**Steps:**

1: **repeat**
2: Perform $\nu_1$ iterations of Relaxation, obtaining $\tilde{u}^{(k)}$:
   i $u^{(k)} := \text{Relax}\left(A, M, f, u^{(k)}\right)$
3: Calculate the residual,
   $r^{(k)} := f - A\tilde{u}^{(k)}$
4: Compute the coarse-grid correction by solving,
   $P^T A P e_c^{(k)} = P^T r^{(k)}$
5: Update the solution with the prolongated coarse-grid correction,
   $\tilde{u}^{(k)} := \tilde{u}^{(k)} + Pe_c^{(k)}$
6: Perform $\nu_2$ iterations of Relaxation, obtaining $u^{(k+1)}$:
   i $\tilde{u}^{(k)} := \text{Relax}\left(A, M, f, \tilde{u}^{(k)}\right)$
7: $k = k + 1$
8: **until** $r^{(k-1)} < \delta$

The two-grid algorithm is described in Algorithm 2.1. The error propagation matrix of steps 4 and 5 of the algorithm is given by

$$\tilde{e}^{(k)} = Ce^{(k)}, \tag{2.22}$$

with $\tilde{e}^{(k)} = u - \tilde{u}^{(k)}$, a definition similar to that of Eq. (2.4), and

$$C = \left(I - P\left(P^T A P\right)^{-1} P^T A\right). \tag{2.23}$$

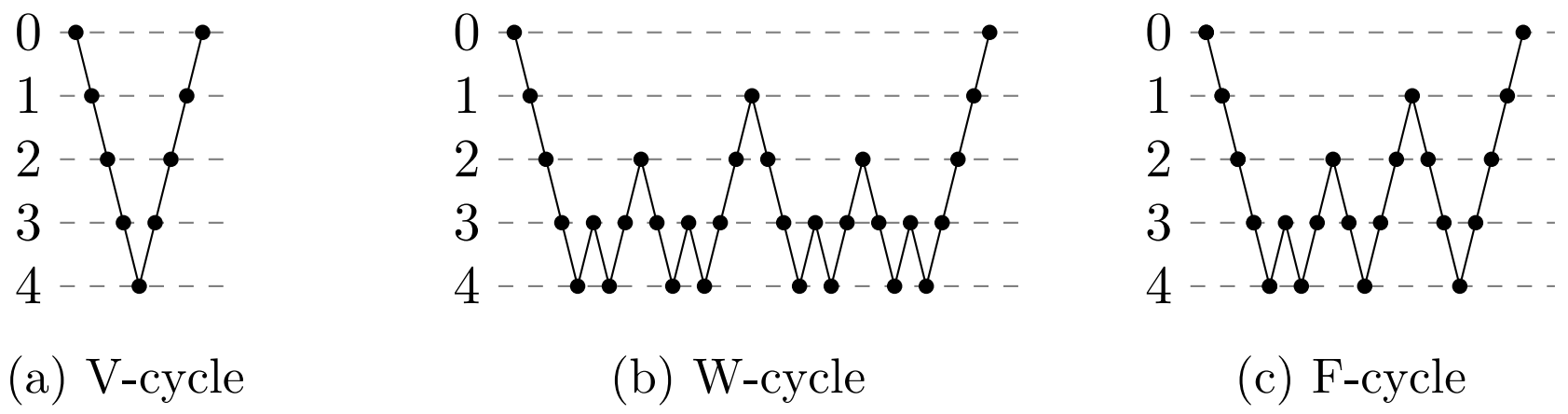

Figure 2.3: Standard types of multigrid cycles. Lines "0" represent the finest grids whereas lines "4" represent the coarsest grids.

Therefore, the error propagation equation of the entire two-grid cycle is given by

$$e^{(k+1)} = \underbrace{S^{\nu_2} C S^{\nu_1}}_{T} e^{(k)}, \tag{2.24}$$

where $S$ is the relaxation error propagation matrix defined in Eq. (2.5) and $\nu_1$ and $\nu_2$ the number of relaxations performed respectively before and after the coarse grid correction. The two-grid cycle is the basis for a family of multigrid algorithms which are implemented, as mentioned earlier, by recursive calls in place of Step 4. As shown in Fig. 2.3, a single recursive call yields the well-known V-cycle, and two recursive calls yield the so-called W-cycle. Another popular choice is the F-cycle, in which a single recursive call to the F-cycle is followed by a call to the V-cycle. The convergence behavior of the two-grid cycle is governed by $T$ in Eq. (2.24), which depends on two components, the prolongation $P$ and the relaxation $S$.

As is the case for matrix S in subsection 2.1, our aim is to minimize the spectral radius of matrix $T$ by selecting optimal nonzero values in the matrix $M$. Without loss of generality, we assume that all the $\nu = \nu_1 + \nu_2$ relaxations are performed before the coarse-grid correction, so $T$ is given by

$$T(M) = (I - P(P^T AP)^{-1} P^T A)(I - MA)^{\nu}. \tag{2.25}$$

As mentioned above, prolongation and restriction operators are required for transferring information from fine grids to coarse grids and vice versa. There are several commonly used restriction operators, including injection, full weighting or Black Box [Den82]. Restriction operators are often chosen to be the transpose of the prolongation (possibly scaled by a constant). Unlike other methods, Black Box is operator-dependent, that is, not based on fixed weighting coefficients in the $P$. This yields robustness of the multigrid solver in the case of fast-varying and even discontinuous coefficients. The exact formulas of calculation of Black Box coefficients can be found in Appendix A.2.

The core part of the present work is related to the relaxation operator (steps 2 and 6 of the two-grid algorithm 2.1). We next describe the common "Red-Black" and "4-Color" strategies for ordering of equations in the relaxation operator. The ordering has a significant effect on the performance of the entire multigrid algorithm.

Multigrid algorithms are used to solve differential equations. The linear equations that approximate these differential equations can be processed sequentially in the relaxation (standard order) or reordered to optimize the outcome (e.g., "Red-Black"). The mathematical description of the reordering of linear equations within Eq. (2.1) is described in Appendix A.3. Linear equations for differential equations also have a spatial interpretation: each equation relates to a location (node) on a grid which discretizes the physical space in which a solution is sought. The different ordering strategies are labeled according to the visual patterns which emerge as the calculation progresses through the grid. On a square grid in 2D, a node has 8 neighbours. The 9 locations (i.e., the neighbours plus the node in the center) form a $3 \times 3$ square called a "stencil". By construction, each row of the matrix $A$ has 9 non-zero values. Alternative stencils can be defined for 2D spaces, such as a 4-neighbour stencil shaped like a plus sign (center plus top, bottom, left and right neighbours) and an "X-shaped" stencil where the neighbours are at the corners of the square which surrounds the central node.

The standard way to order the relaxation is to run through the equations of the system and update the components of the solution by row- or column-major order, resulting in standard Gauss-Seidel relaxation. This standard order is illustrated with a row-major order in Figure 2.4a. All the neighbours that are above the current component and to its left are updated (in green) while others are not (in white). The relaxation is adapted as necessary on the borders of the grid (e.g., in case of periodic boundary conditions, the upper-left corner value does not have left or top updated neighbours). In general, the cyclicity of the PDE influences theses calculations on the grid borders.

Another commonly used ordering strategy is "Red-Black", represented in figure 2.4b which has a checkerboard structure. This strategy is not used in the present research but it is very useful in 2D when the stencil is plus-shaped (center node with top, bottom, right and left neighbours). The ordering consists of first updating all the Red ("1") locations and then updating all the Black ("2") ones. In the case of a plus-shaped stencil, all the calculations with this strategy for each color can be calculated in parallel. The multigrid convergence rate of is also improved with this choice (e.g. [ST82]).

The relaxation ordering that is used in the present work is 4-Color. It differs from Red-Black in that we tile the grid using four colors rather than two. These are ordered as shown in Fig. 2.4c. The relaxation corresponding to the locations with color "2" start only after all the calculations have been completed on the whole grid for locations with color "1", and "3" comes after "2", and finally "4" after "3". The advantage of this ordering is that none of the locations for a given color has a neighbour of the same colour, even on the diagonals. For the stencils with 9 points mentioned earlier, this order makes it possible to calculate and update all the locations for a given colour in parallel. This strategy, like Red-Black, does not require additional calculation time (except for the reordering itself that may be performed in a very effective way), but significantly improves the convergence performance as well as allowing parallel execution of each of the colors.

**Algorithm 2.2** 4-Color Successive over-relaxation

**Input:**

i Discretization of matrix $A$

ii Current solution approximation $u^{(k)}$

iii Right-hand side $f$

iv Relaxation coefficient $\omega$

**Output:**

i Result, approximation after relaxation $u^{(k+1)}$

**Steps:**

1: Perform Weighted Jacobi relaxation of all "1" color nodes,
$u_i^{(k)} := (1-\omega)u_i^{(k)} + \frac{\omega}{a_{ii}}(f_i - \sum_{j\in\{n_i\}} a_{i,j}u_j^{(k)})$

2: Perform Weighted Jacobi relaxation of all "2" color nodes,
$u_i^{(k)} := (1-\omega)u_i^{(k)} + \frac{\omega}{a_{ii}}(f_i - \sum_{j\in\{n_i\}} a_{i,j}u_j^{(k)})$

3: Perform Weighted Jacobi relaxation of all "3" color nodes,
$u_i^{(k)} := (1-\omega)u_i^{(k)} + \frac{\omega}{a_{ii}}(f_i - \sum_{j\in\{n_i\}} a_{i,j}u_j^{(k)})$

4: Perform Weighted Jacobi relaxation of all "4" color nodes,
$u_i^{(k)} := (1-\omega)u_i^{(k)} + \frac{\omega}{a_{ii}}(f_i - \sum_{j\in\{n_i\}} a_{i,j}u_j^{(k)})$

5: Return $u^{(k+1)} := u^{(k)}$

$\{n_i$ refer to neighboring nodes of node $i\}$

As mentioned earlier (subsection 2.1), solution values are updated only after each full iteration in the case of the Jacobi method, unlike in the Gauss-Seidel method where they are updated on the fly (during the iteration). From a conceptual perspective, each of the colors of the Red-Black or 4-Color execution strategies can thus be understood as the implementation of a Jacobi approach to solving the equations for that color because none of the corresponding equations depends on other colors. But taken as whole, the successive calculations for the different colors of the Red-Black or 4-Color strategies yield a reordered Gauss-Seidel relaxation. Applied with a relaxation parameter, it becomes Successive Over-Relaxation (SOR) as shown in Algorithm 2.2. The relaxation parameter does not impair parallelization.

In this work we examine three different relaxations for multigrid. The first is SPAI-0, the simplest version of SPAI in which $M$ is required to be a diagonal matrix [BG02] (Ref. Appendix A.4). The other two are Weighted Jacobi (WJ) and 4-Color Successive over-relaxation (SOR). Choosing optimal parameters (coefficients) for Weighted Jacobi and Successive over-relaxation is crucial for optimizing convergence rates. A common approach in the case of constant (or slowly varying) coefficients uses Fourier analysis, for example [YO98]. The main focus of the present research is to propose an alternative approach to finding the relaxation parameters that are optimal for multigrid.

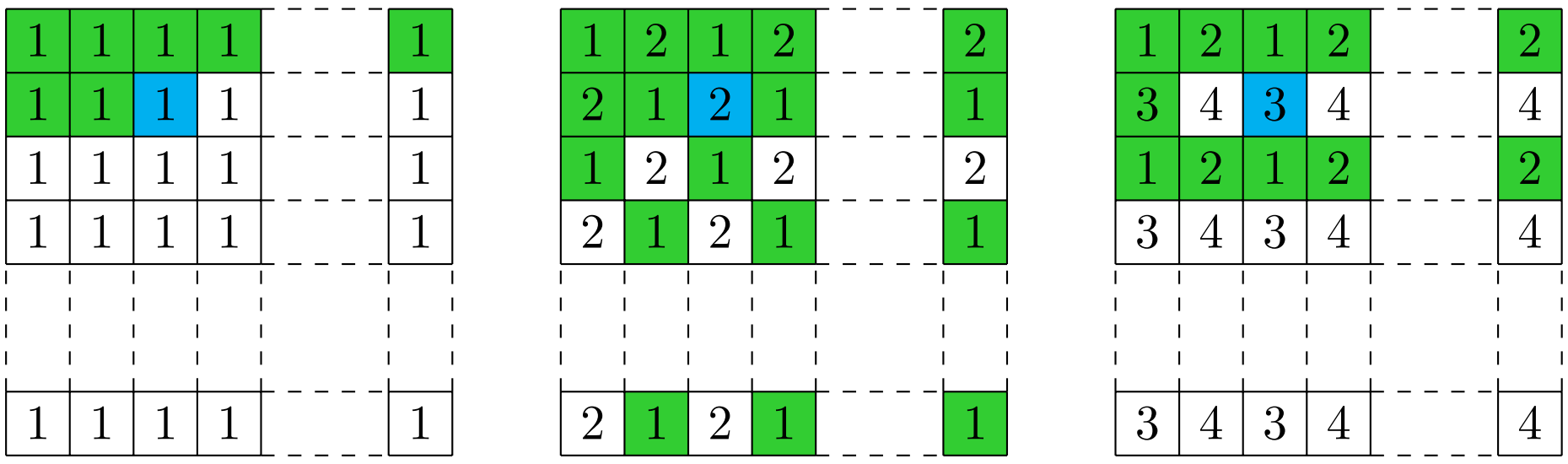

(a) Regular order (b) Red-Black order (c) 4-Color order

Figure 2.4: The blue cells are the current locations at which the relaxation is being calculated. The green cells are the locations at which the relaxation has already taken place. The white cells are the locations at which the relaxation has not yet taken place.

# Chapter 3

# Neural Networks

## 3.1 Introduction to Neural Networks (NN)

How intelligent is Artificial Intelligence? Are Machine Learning and Deep Learning really about learning? And how is all this related to Neural Networks? Since there is no unequivocal agreement about these terms among the researchers (e.g. [Kor97]), some disambiguation is needed. First, "Neural Networks" shall stand for "Artificial Neural Networks" (as opposed to those in our brains). Deep Learning (DL), is a sub-discipline of Neural Networks (NN) which, in turn, is a sub-discipline of Machine Learning (ML), all branches of Artificial Intelligence (AI), defined by a system's ability to perform tasks without having been explicitly programmed to do so.

Machine Learning is a concept in which a system improves its performance through some 'learning process', usually achieved via iterated interaction with a large data set. Classical examples of ML algorithms are: Linear Regression, Support Vector Machine (SVM), and Evolution Algorithms. To contrast ML from general AI systems, consider SHRDLU [Win71]—a computer program communicating with its user and executing basic commands in virtual space using standard English language commands. SHRDLU does not undergo training and its 'intelligence' is fully programmed into it.

As mentioned, Neural Networks (NN) are a subset of ML. What, then, has recently made it so popular, that in common language artificial intelligence and machine learning have become almost synonymous to NN? The answer is probably the flexibility of NN, enabling it to tackle many AI tasks. Theoretically, neural networks can imitate any general recursive function ("computable" function), but in practice the implementation may be a challenging task and, in some cases, verge on the impossible[1]. Like any classical algorithm, a NN may have different sizes and types of inputs (images, text, sequences of sensor data, etc.) or no input at all. The outputs of NNs also do not have any specific limitations. The only practical limitations are hardware availability

[1] "Neural networks are theoretically capable of learning any mathematical function with sufficient training data, and some variants like recurrent neural networks are known to be Turing complete." Quoted from [Agg18], page VII

and the training time. The commonly used name for a NN with many hidden layers (explained below) is Deep Learning (DL) or Deep NN. Some recent deep NNs have dozens of billions of parameters (i.e., trainable neurons, for example, GPT-3, MEB). The downsides of such NNs are their difficult training which requires huge amounts of training data, their relatively slow execution process, and their requirement for high-end computational hardware.

There is a huge variety of NN structures. Choosing a particular structure is artwork in part and, as such, there are not always right and wrong solutions. In what follows, fully connected networks are considered in which all neurons from a layer are connected to all the neurons of the previous layer (as well as to all the neurons on the next layer). A given neuron (Fig. 3.1) maintains a separate floating point value (weight) for each of the neurons of the previous layer. In execution mode, this neuron performs two steps. At first, it multiplies each output of the previous layer by its respective weight and sums over the results. The result of this first step constitutes the input to the activation function of the neuron which gives the output value of the neuron. Activation functions usually bound the output values and add some non-linearity to the process, thereby increasing the range of functions that the network is able to imitate. This process is repeated for each neuron of each layer and for each layer of the network from first to last. There is no theoretical limitation to the number of neurons in a layer, nor to the number of layers in a neural network. Although only a small variate of neuron activation functions is being used in practice, it is by no means exhaustive.

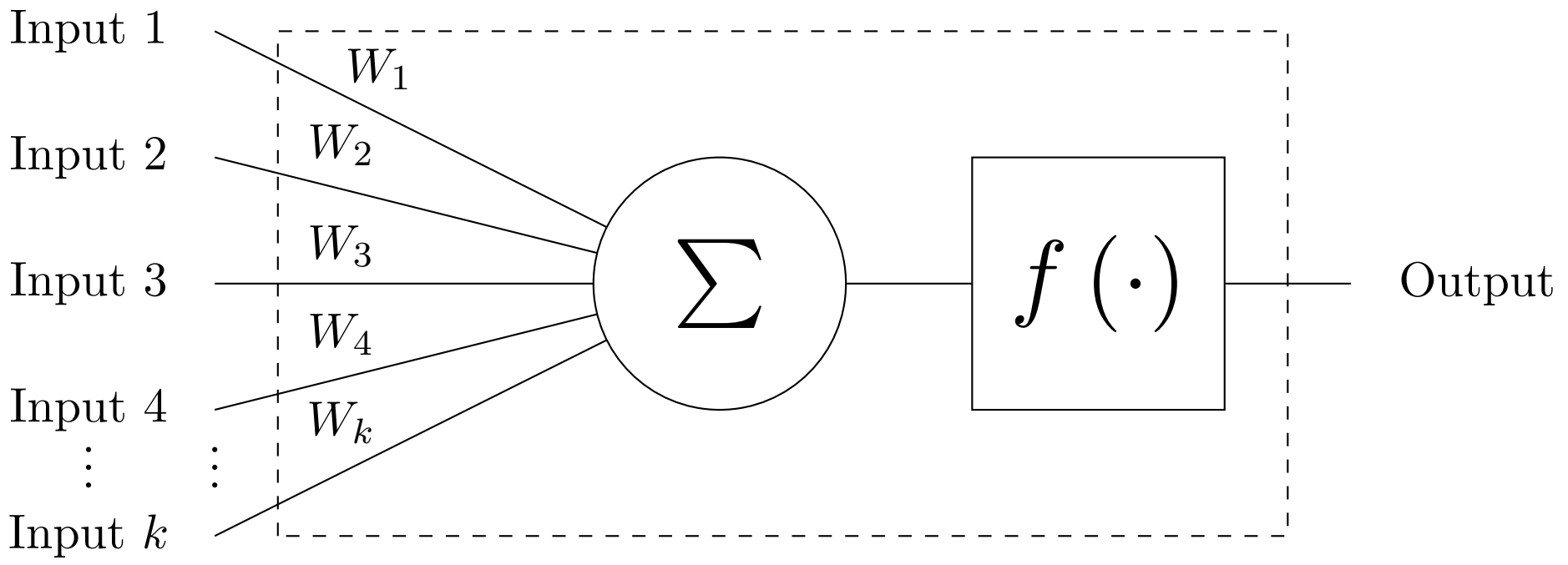

Figure 3.1: Structure of a single neuron (dashed box). Weights $W_k$ are associated to the inputs. The weighted inputs are summed $\sum$. This sum is an input to the neuron's activation function $f(\cdot)$, which returns the final output of the neuron.

The fully connected neural network described above is one of the simplest architectures (cf. Fig. 3.2 for an example of a NN with three hidden layers). Among other commonly used architectures are Convolutional Neural Networks (CNN), Long Short-Term Memory (LSTM, a subset of Recurrent Neural Networks, RNN) and Autoencoder. Covering those would go beyond the scope of the current work. Common to all neural networks is a loss function, encoding the "direction" in which the weights should be changed as part of the learning process, explained in the next paragraph. In addition, pre- and post-processing functions can be added to a network. For example, the data input at the entrance of the neural network (first layer) can be raw data or a filtered version of the same data. Finally, a limitation on the design of all networks must be mentioned: all the functions used between the trainable neurons and the loss function of the network must be differentiable functions, otherwise the learning process would not be possible.

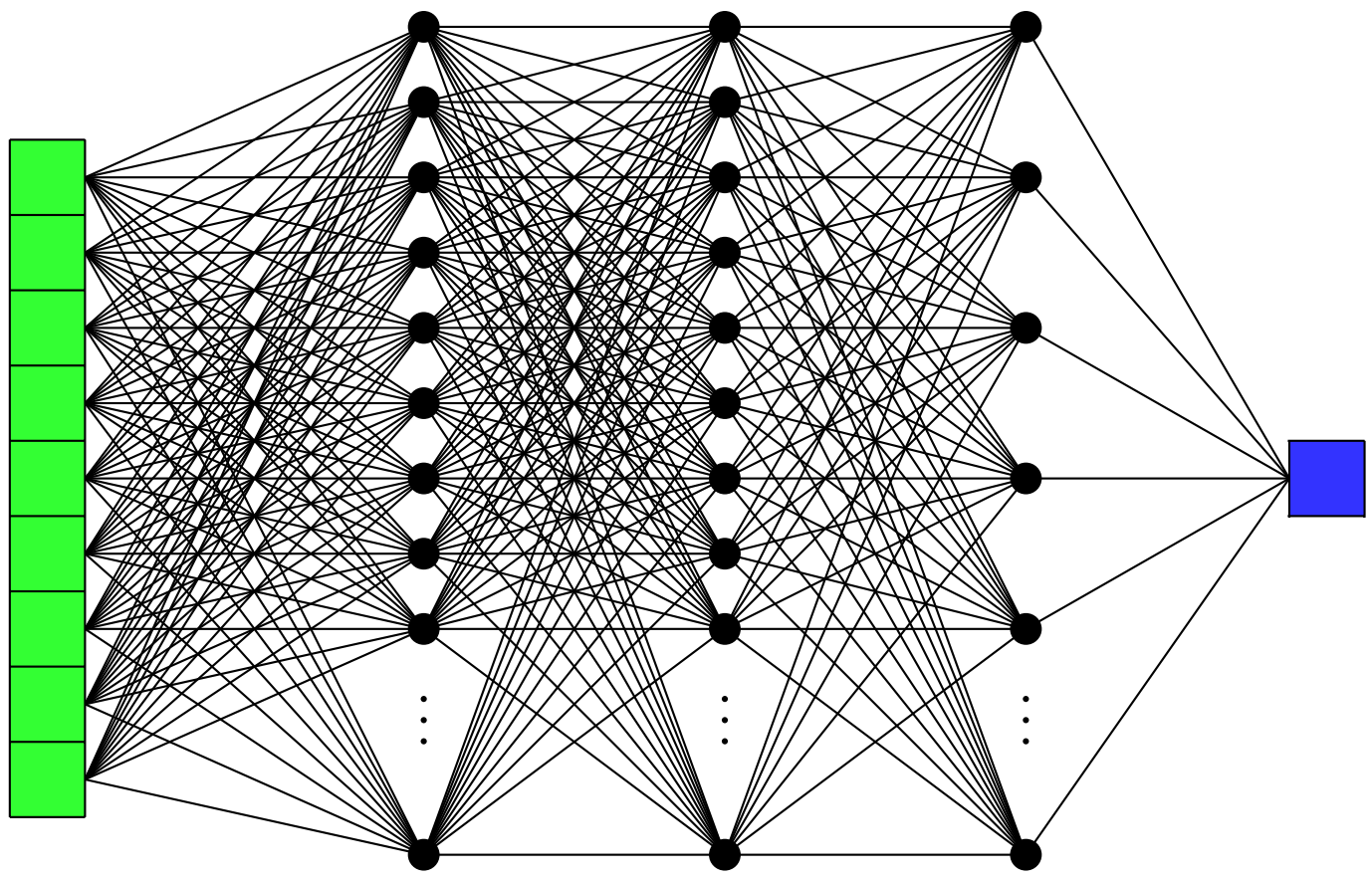

Figure 3.2: Structure of a Neural Network. Inputs are in green, followed by three hidden layers and by the output in blue.

Executing a trained neural network is nothing but applying a list of given functions to some data. The training session of a NN is where the 'magic' happens. Although a single change of a neuron during a training cycle is usually negligible in terms of network performance, repeated changes gradually imprint meaningful changes in network behaviour. Initially, weights are assigned randomly throughout the network (initialisation phase). As the training is executed, these weights progressively lead the network to "learn" patterns. NN learning is commonly categorized by two types: supervised and unsupervised. While in supervised training the desired outcome is part of the training data, unsupervised training is performed without an explicit definition of the desired outcome. Some examples of unsupervised learning are "anomaly detection" and "clustering". In both cases, the core part of the process lies in the loss function. While in supervised learning this function is natural, defined by minimizing the distance to the desired outcome (e.g., Euclidean norm of difference between current output and the desired outcome), in unsupervised learning the function reflects the learning hypothesis, similarly to the way it works in optimization problems. In this case the loss function should be built carefully to avoid divergence.

The learning (training) process utilizes *backpropagation*, which is an efficient implementation of gradient descent. As is well-known, for any scalar field, differentiable functions have a vector value at each point that defines the direction of the gradient (fastest ascent direction). Moving in the direction opposite to that given by the gradient leads to the lowest point around the current location (steepest descent). Choosing an appropriate step size and its rate of decay during the progression of the learning process is a separate topic of neural network training. A detailed explanation of the subject goes beyond the scope of this work, but the guideline is: if the step is too small, or the decay rate too fast, the process may not reach a minimum; On the other hand, if the step is too big it might overshoot without converging to the minimum. A proper choice should decrease the step size as the result approaches the minimum so as not to skip over it. Due to the random initialisation of weights and random runs through the training data (among other factors) the minima that are found during training usually return function values that are very close to the global minimum[2]. The training (back-propagation of the loss function in the direction of gradient descent) is the reason why all the functions used between trainable neurons and the loss function of the network must be differentiable as mentioned earlier.

## 3.2 Neural Networks for Multigrid relaxation

Among the many uses of neural networks, their application to multigrid relaxation presented in chapter 2 is promising. If neural networks did not have structural limitations, using the multigrid convergence rate as a loss function for the training process would make it easy to optimize the relaxation component of the multigrid. However, aforementioned limitations prevent such straightforward implementation of neural networks. Two major issues are the practical impossibility to calculate derivatives in the case of multigrid and the fact that the convergence rate function is not differentiable and therefore cannot be used as the loss function. First, the recursiveness of multigrid (mentioned in subsection 2.3) makes it inapplicable to calculate the derivatives of all the relaxation parameters for most problems except very small ones. This issue is solved by training the NN with the Two-grid method instead of the full multigrid algorithm. The Two-grid method has an explicitly differentiable formula (Eq. (2.25)). But this solution comes with another constraint. The limitation of the Two-grid method is the necessity to solve the coarse grid problem exactly. This limits the maximum size of the grid on which it is possible to train the NN. Secondly, the asymptotic convergence rate of multigrid is given by the spectral radius $\rho(T)$ of matrix $T$ (Eq. (2.25)). A direct approach to minimizing the spectral radius of $T$ is problematic because the spectral radius is a non-smooth function of the nonzero values of $M$ [KDO17]. Hence, training the NN on the loss function of the spectral radius of $T(M)$ and backpropagating it is impossible. Following [KDO17], Gelfand's formula offers a way out, relating the spectral radius of a matrix to the Frobenius norm of powers of the matrix:

[2] "Local minima are problematic only when their objective function values are significantly larger than that of the global minimum. In practice, however, this does not seem to be the case in neural networks. Many research results have shown that the local minima of real-life networks have very similar objective function values to the global minimum. As a result, their presence does not seem to cause as strong a problem as usually thought." from [Agg18], page 152

$$\rho(T) = \lim_{\alpha \to \infty} \|T^\alpha\|^{\frac{1}{\alpha}}. \tag{3.1}$$

The solutions to these two issues that constrain the application of neural networks to the relaxation of multigrid will be presented in detail in Chapter 4.

# Chapter 4

# Numerical Experiments

## 4.1 Introduction

In this chapter we describe the experiments that were executed and their results chronologically. The general structure of each phase of our research is as follows: rationale, description, training setup, testing setup, results, conclusion.

This work focuses on the use of NNs for the relaxation of multigrid algorithms. It explores the effectiveness of NN for the convergence to solutions to the periodic diffusion partial differential equation (2.14). Two types of restrictions and prolongations are used: Bilinear and Black Box (which is mentioned in section 2.3). The NN training process is executed on randomly generated artificial data. For testing, other data sets are generated in the same way. The exact parameters used for each experiment are given in each of the corresponding sections below.

As already mentioned, training the NN on the loss function of the spectral radius of $T(M)$ and backpropagating it is practically impossible because the spectral radius is a nonsmooth function of the nonzero values of $M$. The loss function that is chosen for this work is $\|T^\alpha\|$. The Original Gelfand's formula (3.1) requires $\sqrt[\alpha]{\cdot}$, however, for our needs of minimization this operation can be avoided because any $\alpha$-th root function is a monotonically non-decreasing function for any positive values, therefore:

$$\arg\min_M(\rho(T(M))) = \arg\min_M(\lim_{\alpha\to\infty} \|T(M)^\alpha\|). \tag{4.1}$$

In other words, for big enough values of $\alpha$, approximately the same $M$ minimizes both of the functions. Gelfand's formula is valid for any norm. Among all norms, the Frobenius norm is chosen because it is differentiable and thus allows backpropagation. Finding the range of $\alpha$ values that fit the problem can be done by trial and error. Values of $\alpha$ that are too small lead to poor approximations, and values of $\alpha$ that are too big make the training process slow due to the necessity to execute complex differentiations as part of the backpropagation process. In practice, we found that $\alpha$ values in the 10 to 40 range are a good compromise, but even smaller values yield reasonable results.

## 4.2 Stencil-based NN

Although the experiments described in this section did not yield the desired results, they led to important considerations and set the conceptual stage for section 4.3, the main scientific contribution of the present work. One of the well-known methods to provide the approximate inverse matrix $M$ of matrix $A$ in Eq. (2.2) is SPAI-0 (cf. Appendix A.4). This part of the research examines the use of NN as an alternative to SPAI-0 and Weighted Jacobi for this task. The NN chosen for this mission should be a narrow-structured network to allow for fast and computationally cheap execution processes. This NN (as SPAI-0 does) uses as input the 9-value stencil shown in Fig. 2.2a (for the diffusion problem, it is in fact sufficient to input the four values of $g$ as explained in detail in section 2.2 but we prefer to maintain generality for other nine-point operators). Let $a_{p,i}, \dots, a_{q,i}$ be these 9 non-zero values of row $i$ of matrix $A$. Like SPAI-0, the NN returns a single value which corresponds to the diagonal element of the matrix $M$ associated with a given row of $A$. The NN is deemed superior when, during the testing phase, the convergence rate of the entire method which uses the NN for relaxation outperforms the convergence rate of the entire method which uses SPAI-0 (or Weighted Jacobi) for the relaxation. The entire method we use Two-grid for testing in the first experiment (section 4.2.1) and Multigrid in the remainder.

The NN is built with three fully-connected hidden layers presented in Fig. 3.2 and with the ReLU activation function. The number of neurons in the first, second and third layers is 36, 36 and 18, respectively. As a NN optimizer we use the standard stochastic gradient descent. The type of restriction and prolongation used here is Bilinear (Fig. 4.1).

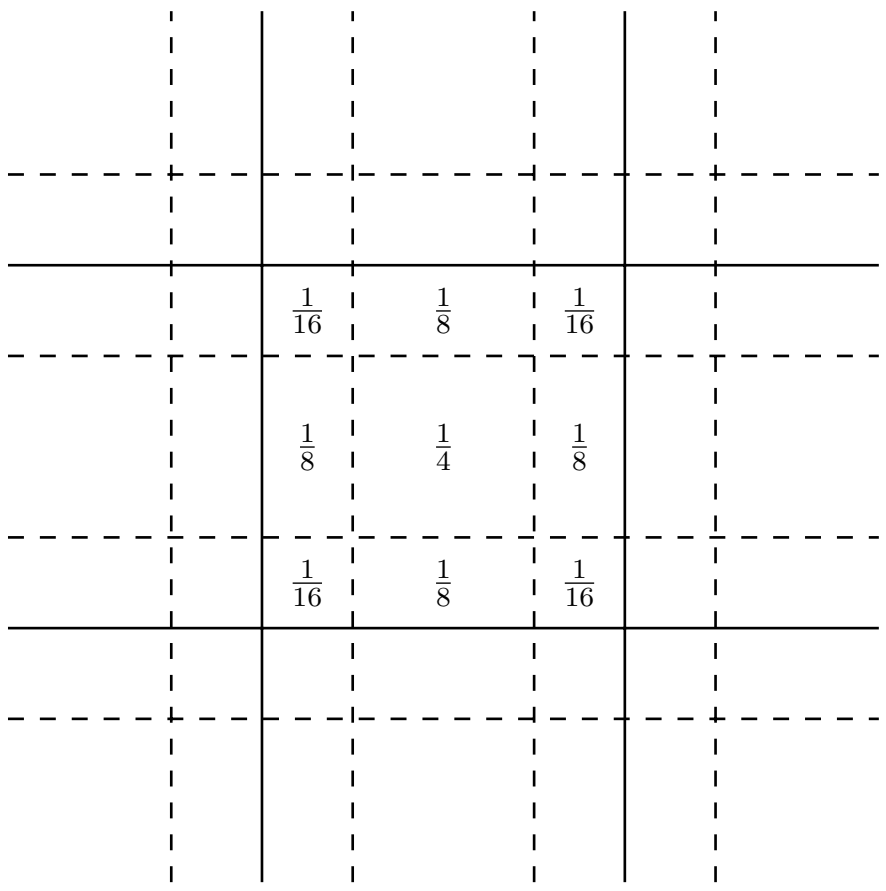

Figure 4.1: Solid mesh - coarse grid, Dashed mesh - fine grid. Coarse to fine grid interpolation with coefficients.

Table 4.1: Training parameters

| Parameters | |
|---|---|
| Method | Two-grid |
| grid size | 16 × 16 |
| data distribution | log normal |
| number of samples in batch | 20 |
| training / validation division | 0.8 / 0.2 |
| epochs between batch regeneration | 200 |
| loss function | mean square |
| $\alpha$ | 10 |
| $\delta$ | 0.01 |
| $\nu$ | 1 |
| Prolongation | Bilinear |

### 4.2.1 NN vs. SPAI-0

We first test whether the NN can outperform the SPAI-0 relaxation in the multigrid algorithm.

The Sparse Approximate Inverse (SPAI) method [BG02] defines a sparse approximate inverse $M$ of a given invertible matrix $A$ by minimizing the Frobenius norm of $I - MA$, subject to a prescribed (or dynamically determined) sparsity pattern of $M$. SPAI has been shown to yield excellent relaxation operators for multigrid solvers, so beating SPAI in its own game is challenging. The SPAI-0 variant of SPAI constrains $M$ to be a diagonal matrix. This way, each of the values $m_k$ on the diagonal of $M$ is influenced by only one corresponding row of matrix $A$,

$$m_k = \frac{a_{kk}}{\sum_{i=1,\dots,n} a_{ki}^2} = \frac{a_{kk}}{\sum_{i=p,\dots,q} a_{ki}^2}. \tag{4.2}$$

The full proof of why Eq. (4.2) minimizes the Frobenius norm can be found in Appendix A.4.

To train the NN, a Two-grid method is implemented (using TensorFlow, a python-based software library for machine learning and artificial intelligence). The training is executed with the parameters that are listed in table 4.1. The NN is trained with data generated using a log normal random distribution. The size of the matrix **g** for training is 16x16. The NN is trained to minimize $\|T^\alpha\|_F$, the Frobenius norm of $T$ to the power $\alpha$. The $\delta$ parameter represents the value that is added to the diagonal of $A$ as described in section 2.2, to make $A$ diagonally dominant (thus preventing it from being singular).

Every few cycles, the NN is saved (by a snapshot of all the weights of the NN). This snapshot is used to evaluate the progress of the training. However, it can be seen that the training process is unstable. As shown on Fig. 4.2, the training and validation losses do not converge even after a large number of training epochs ($4.75e6$).

The testing process is similar to the training process and the parameters are listed in table 4.3. The main differences are that the testing is run on a set of data which is not regenerated during the run, and the testing also calculates the spectral radius of matrix $T$ (not calculated during the training process).

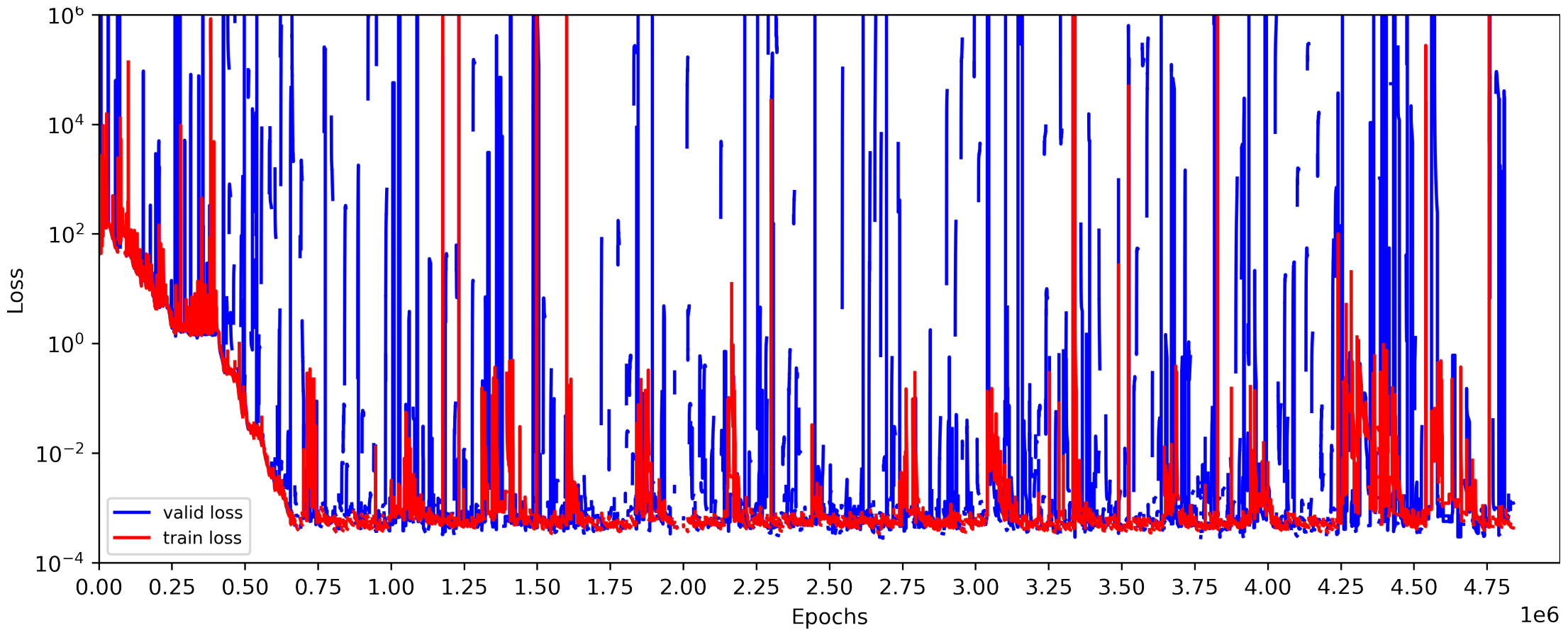

Figure 4.2: Training process. Red is the loss of the training dataset as a function of the number of epochs (in millions). Blue is the loss of the validation dataset.

Table 4.2: Results of Gelfand's formula and spectral radii for NN and SPAI-0 relaxations within Two-grid using the parameters of Table 4.4. The NN was trained with parameters of Table 4.1. "Std." stands for Standard deviation.

| | Gelfand's formula | Spectral Radius |
|---|---|---|
| Method | Mean ± Std. | Mean ± Std. |
| NN | $0.636 \pm 0.07$ | $0.588 \pm 0.07$ |
| SPAI-0 | $0.682 \pm 0.055$ | $0.652 \pm 0.061$ |

A straightforward measure for comparison between NN and SPAI-0 is $\|T^\alpha\|^{\frac{1}{\alpha}}$ (Gelfand) with $\alpha = 10$. The results for spectral radius are comparable. Table 4.2 shows that the NN outperforms SPAI-0 by 0.046 on average. When the spectral radius is calculated instead of Gelfand, the picture is similar (same table). The spectral radii are improved on average by 0.064 over SPAI-0. A deeper investigation can be done with Figure 4.3, which shows histograms of the differences between the values of each of the 10000 examples. These histograms show that not only does the NN outperform SPAI-0 on average but, as one might expect, it also vastly outperforms SPAI-0 in the percentage of samples (94.82% for Gelfand's formula and 96.35% for the Spectral Radius). However, these histograms have longer tails from the side of values that are greater than the average (when the NN is relatively less effective). The shape of the corresponding density maps of the spectral radii of the NN versus SPAI-0 is shown Fig 4.4. As we already know, most of the samples are under the diagonal (spectral radius given by the NN smaller than by SPAI-0). But an interesting observation is that some samples have a NN spectral radius bigger than 1. In other words, these samples do not converge for NN-based relaxation. As can be seen in Fig. 4.5, the common feature of all samples for which $\rho(T) > 1$ is that they have at least one relatively high value on the diagonal (above 50). This suggests that the input data to the NN should be normalized (section 4.2.2).

Table 4.3: Test parameters

| Parameters | |
|---|---|
| Method | Two-grid |
| grid size | 16 × 16 |
| data distribution | log normal |
| number of samples | 10000 |
| $\alpha$ | 10 |
| $\delta$ | 0.01 |
| $\nu$ | 1 |
| Prolongation | Bilinear |

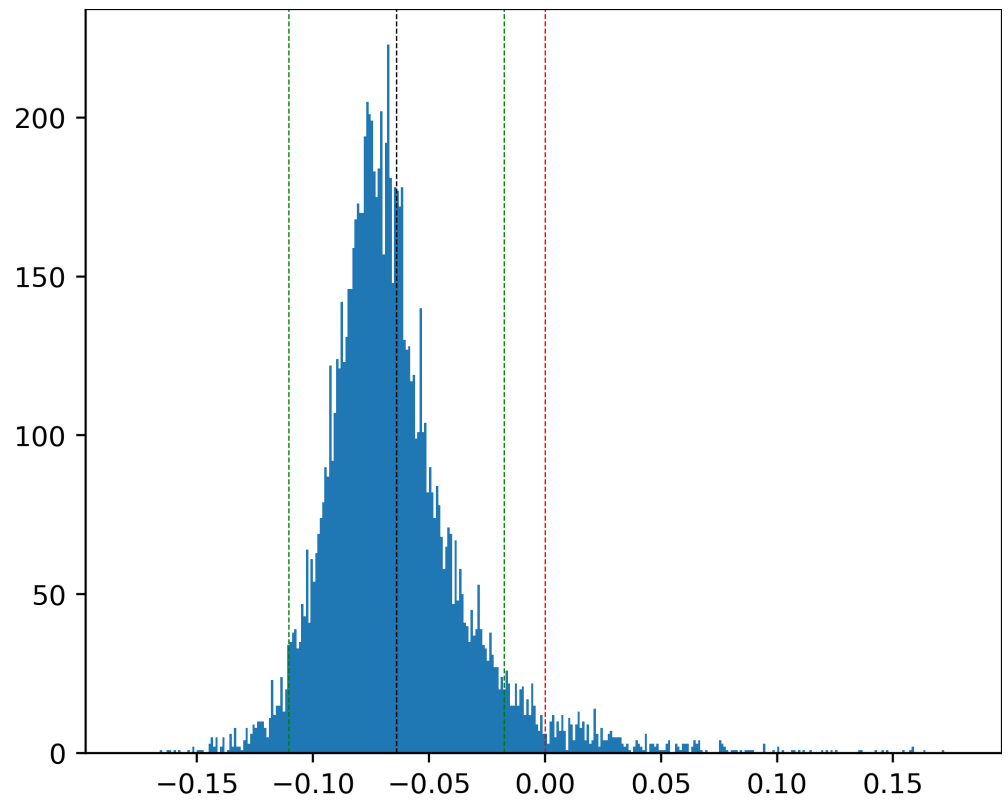

(a) Spectral Radius

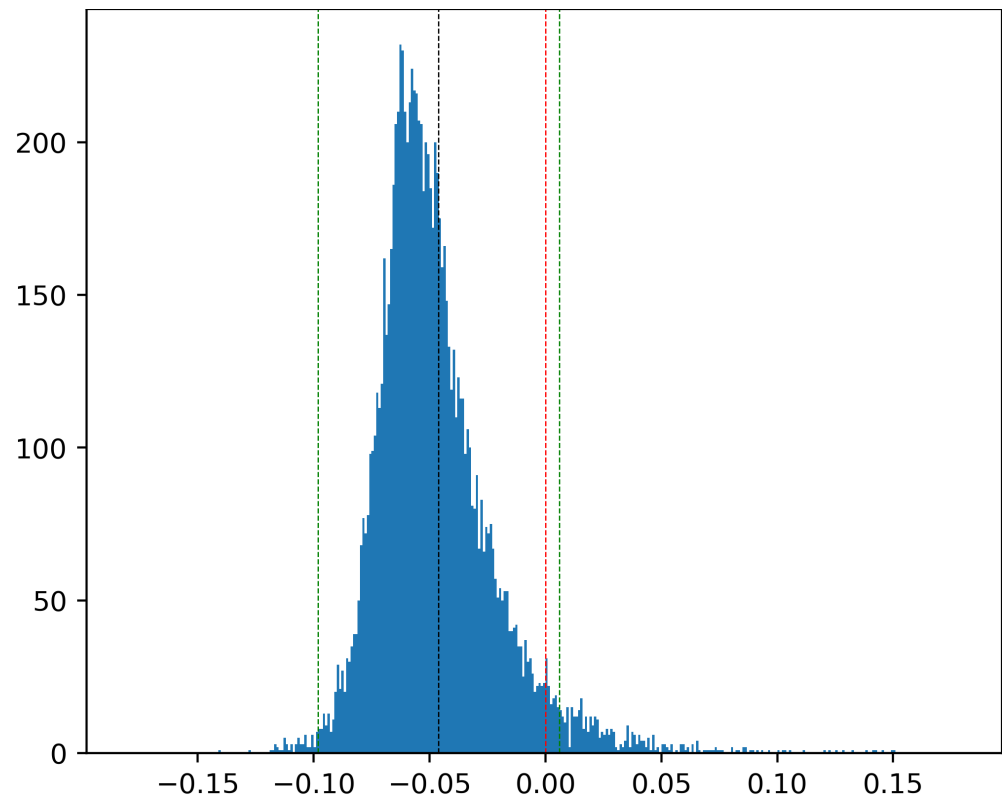

(b) Gelfand's formula

Figure 4.3: Histograms of comparison of results given by the NN and SPAI-0 for ten thousand data points. Histogram (a) plots the differences between spectral radii given by the NN and spectral radii given by SPAI-0. Smaller radii for the NN than for SPAI-0 are located left of the red line. Histogram (b) plots the differences between the approximations of spectral radii given by Gelfand's formula ($\alpha = 10$) for the NN and for SPAI-0. The vertical black line is the average. Green lines indicate $\pm 1$ standard deviation. Red lines correspond to equal results for the NN and SPAI-0.

### 4.2.2 Normalized NN vs. SPAI-0

As seen in the previous section, if the input data of the NN is not normalized, the training takes a long time and shows instabilities (the loss going down and back up erratically). One common practice of data preprocessing for Neural Networks is *Feature Normalization*, which "often does ensure better performance, because it is common for the relative values of features to vary by more than an order of magnitude."[1]. Normalization may eliminate the adverse effect of outliers cf. previous section.

[1] from [Agg18], section 3.3.2 Feature Preprocessing

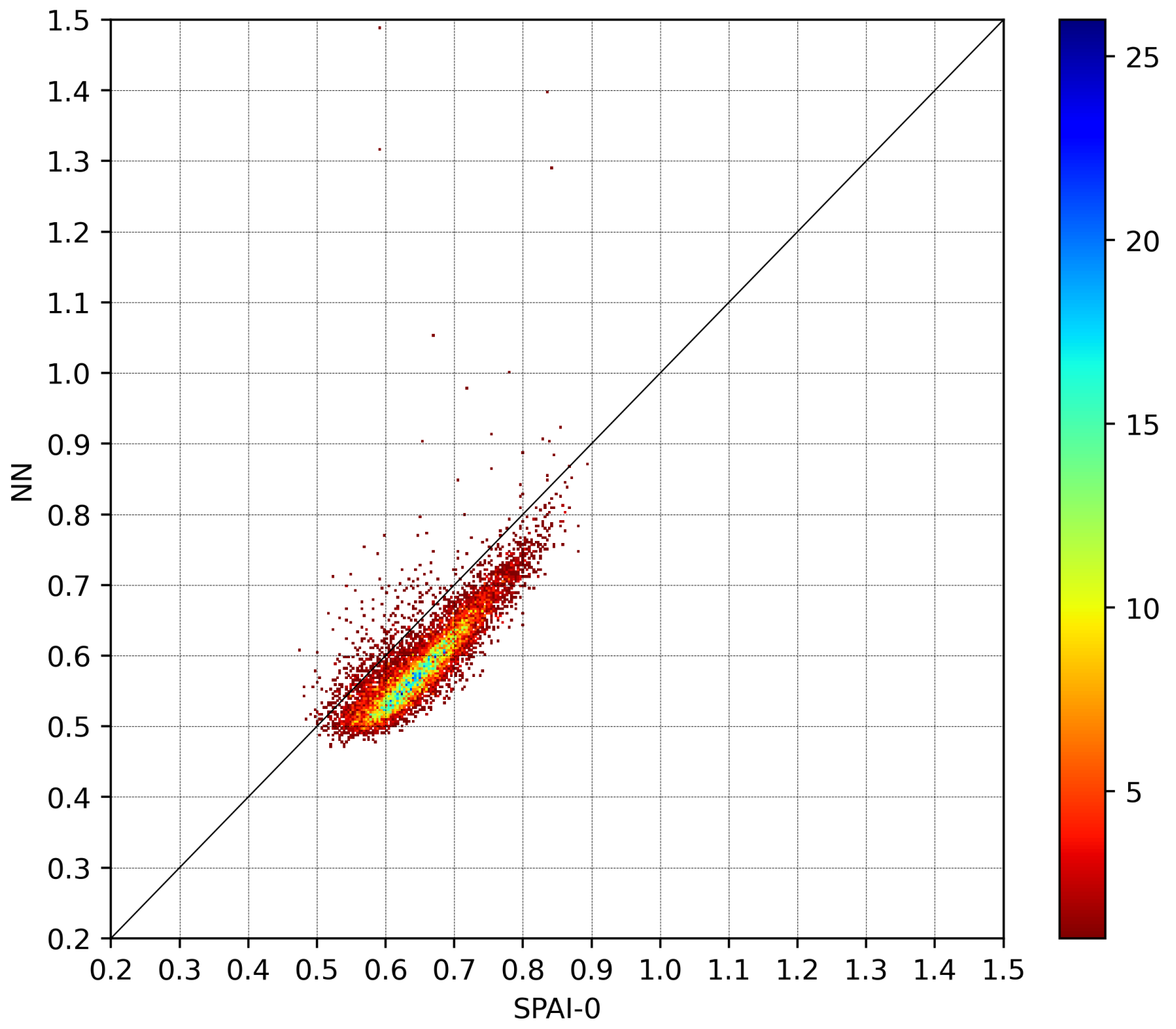

Figure 4.4: 2D histogram of spectral radii of NN versus SPAI-0. The diagonal corresponds to samples leading to equal spectral radii for NN and for SPAI-0.

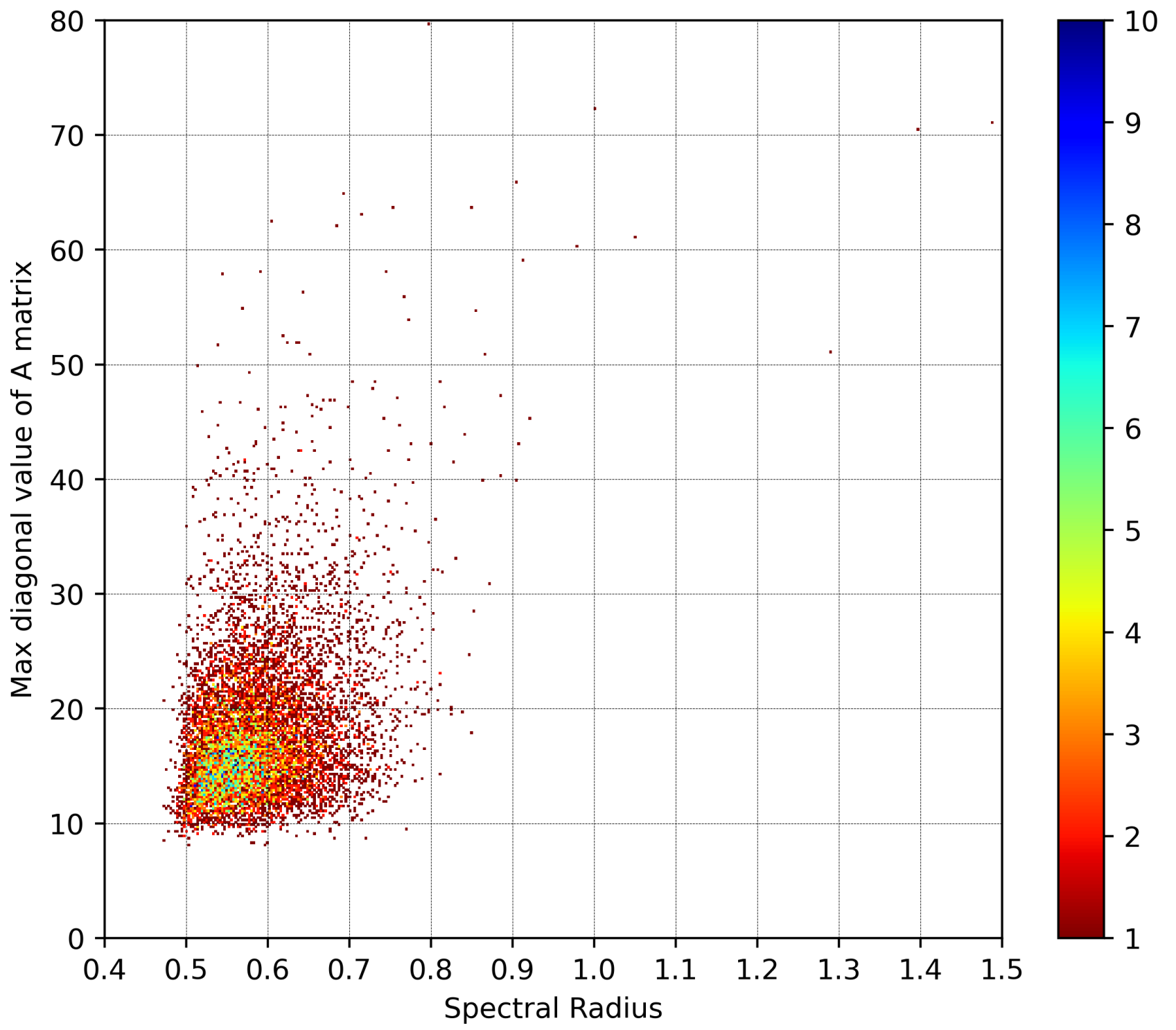

Figure 4.5: Density map of the maximum diagonal value of $A$ matrices versus the convergence rate of $T$ matrices. Large spectral radii of the $T$ matrix are observed only when diagonals of $A$ contain at least one large value (top right of the chart).

Table 4.4: Training parameters

| Parameters | |
|---|---|
| Method | Two-grid |
| grid size | 16 × 16 |
| data distribution | log normal |
| number of samples in batch | 20 |
| training / validation division | 0.8 / 0.2 |
| epochs between batch regeneration | 200 |
| loss function | mean square |
| $\alpha$ | 10 |
| $\delta$ | 0.0001 |
| $\nu$ | 1 |
| Prolongation | Bilinear |
| Input values | Normalised |

In the case of stencil-based inputs to the NN which emulates the SPAI-0 method, the obvious way to implement the normalization is by dividing the values of each row of $A$ by the value on the diagonal of that row. The implementation of this normalization is straightforward. From Eq. (4.2),

$$m_k = \frac{a_{kk}}{\sum_{i=p,\dots,q} a_{ki}^2} = \frac{1}{a_{kk}} \cdot \frac{1}{\sum_{i=p,\dots,q} (a_{ki}/a_{kk})^2}. \tag{4.3}$$

The NN is implemented in the following way:

$$m_{k_{not_norm}} = f_{NN}(a_{kp}, \cdots, a_{kq}) \Longrightarrow m_{k_{norm}} = \frac{1}{a_{kk}} \cdot f'_{NN}(a_{kp}/a_{kk}, \dots, a_{kq}/a_{kk}), \tag{4.4}$$

where $f_{NN}$ represents the NN function trained with un-normalized data and $f'_{NN}$ represents the NN function trained with normalized data. $m_{k_{not_norm}}$ and $m_{k_{norm}}$ are the diagonal values calculated by $f_{NN}$ and $f'_{NN}$ respectively. As a result of normalization, values input to the NN function are in $[-1, 1]$.

As mentioned earlier, the main goal of the current experiment is to test the training process with normalized data. Table 4.4 presents the training parameters. In addition to the normalization of data, the value of $\delta$ was decreased by two orders of magnitude (in practice, this value is large enough to ensure the non-singularity of matrix $A$).

As can be seen from Figure 4.6, the training process is improved. There are fewer jumps in the loss values (the existing jumps can be explained by changes of data sets).

The duration of training is also shortened by an order of magnitude, which allows additional training runs with different parameters. The same training process was thus executed with $\alpha = 1, 2, 4, 6, 8, 10$ and $\nu = 1, 2, 3, 4$ (24 different training runs in total). Additionally, the NN validation loss curve (after the first part of training) shows a high degree of correlation with the loss curve for SPAI-0. This means that samples that are relatively easy for SPAI-0 are similarly easy for the NN and vice versa.

To simulate a realistic case, the test is performed with the Multigrid method. A significant change compared to the testing process of section 4.2.1 is the size of the grid. In order to examine the robustness on bigger grids, the test is executed on a $64 \times 64$ grid. In addition to the configuration specified in Table 4.5, the tests were performed with the following numbers of Pre- and Post-Relaxations, $\nu_1$ and $\nu_2$ respectively : $\{1, 0\}$, $\{0, 1\}$, $\{1, 1\}$, $\{2, 1\}$. In order to keep this section focused, the only test results that are presented here are for $\nu_1 = 1, \nu_2 = 0$ and $\nu_1 = 2, \nu_2 = 1$. Additional results can be found in Appendix A.5.

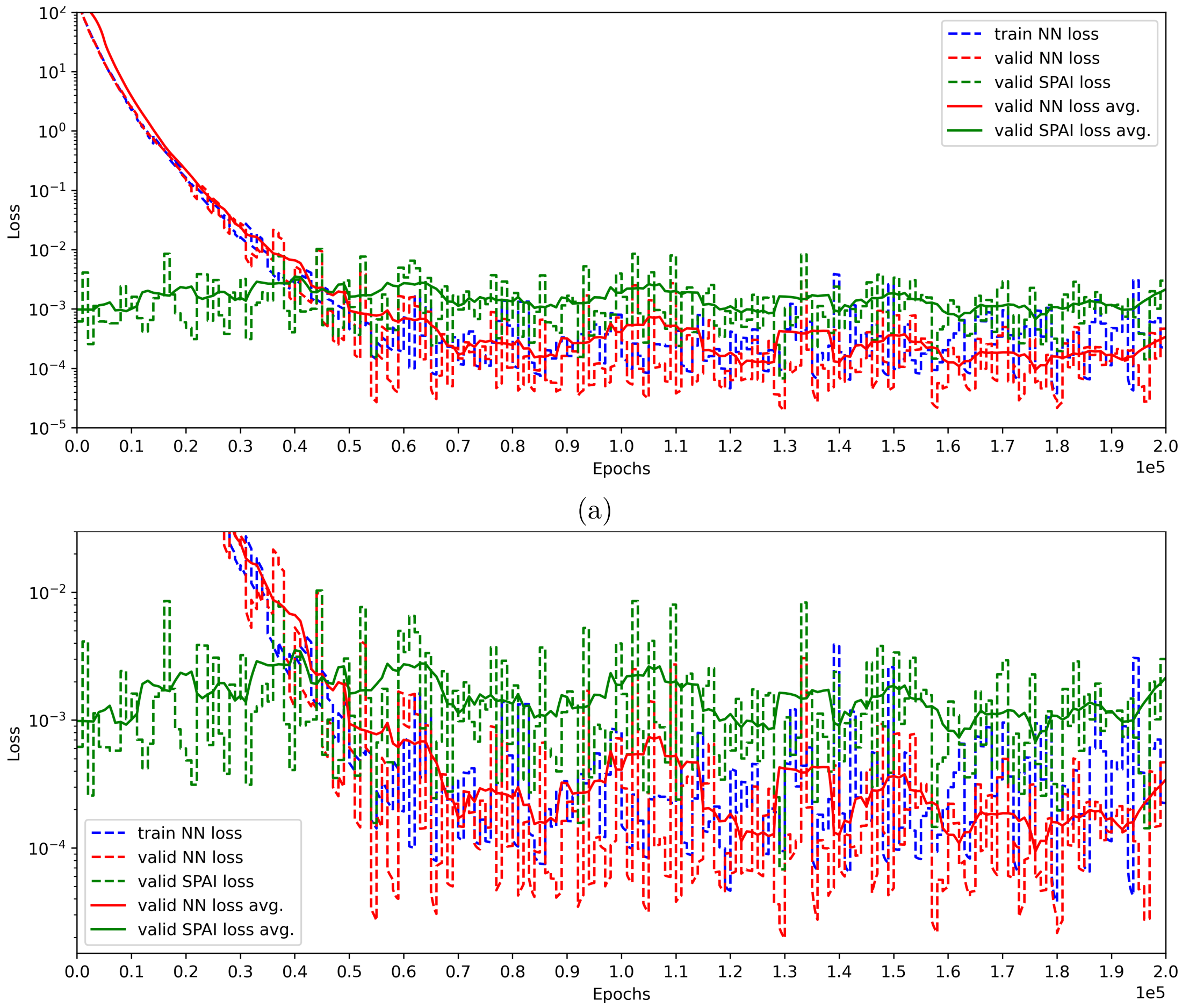

(a)

(b)

Figure 4.6: Plots (a) and (b) represent the loss functions corresponding to the training and validation of the NN (respectively blue and red dashed lines), and to the validation implemented with SPAI-0 (green dashed line). The (x-axis) represents the number of epochs. The (y-axes) plot the training losses. Figure (b) is a close-up of the curves of figure (a). The same validation data sets were used for both NN and SPAI-0. The $1e4$ average values are represented by solid curves in the same colors.

Table 4.5: Test parameters

| Parameters | |
|---|---|
| Method | Multigrid |
| Type | V-cycle |
| grid size | 64 × 64 |
| data distribution | log normal |
| number of samples | 200 |
| $\delta$ | 0 |
| Pre-Relaxation | 1 |
| Post-Relaxation | 0 |
| Prolongation | Bilinear |
| Input values | Normalised |

Figure 4.7 shows residuals as functions of multigrid cycles during the testing part. Because the graph is semi-logarithmic, the value that represents the average with greater relevance is the geometric mean. Based on this measure, the average convergence rate was calculated in an area where the values have a smooth slope (15 to 40 cycles, see table 4.6). During the test stage, each of the test execute we test SPAI-0 relaxtion as well as with each of trained NNs (As mentioned earlier, the NNs are different in the parameters with which they were trained). As can be learned from the table, the NN trained on normalized data outperforms SPAI-0 in most of the training cases (percentage of NN wins) and in average residual convergence rates. For example, The best convergence rate for $\nu_1 = 2, \nu_2 = 1$ test is for a training run with $\nu = 3$ (which makes sense with the assumption from section 2.3, $\nu = \nu_1 + \nu_2$) and $\alpha = 10$ (bigger $\alpha$ improve the approximation of Gelfand's formula). Other results are difficult to explain and require in-depth analysis. For example, it is not immediately clear why the best convergence rate for $\nu_1 = 1, \nu_2 = 0$ is obtained when $\nu = 3$. Nor is it clear why, for the same test, the convergence rate for $\nu = 2$ is worse than for $\nu = 1$ and $\nu = 3$. Other points are left pending as well, like the reason why $\alpha = 4$ minimizes the convergence rate when $\nu = 3$ (contrary to what was expected, increasing of $\alpha$ was expected to improve the convergence rate).

The following section focuses on the comparison of NN-based relaxation with weighted Jacobi-based relaxation.

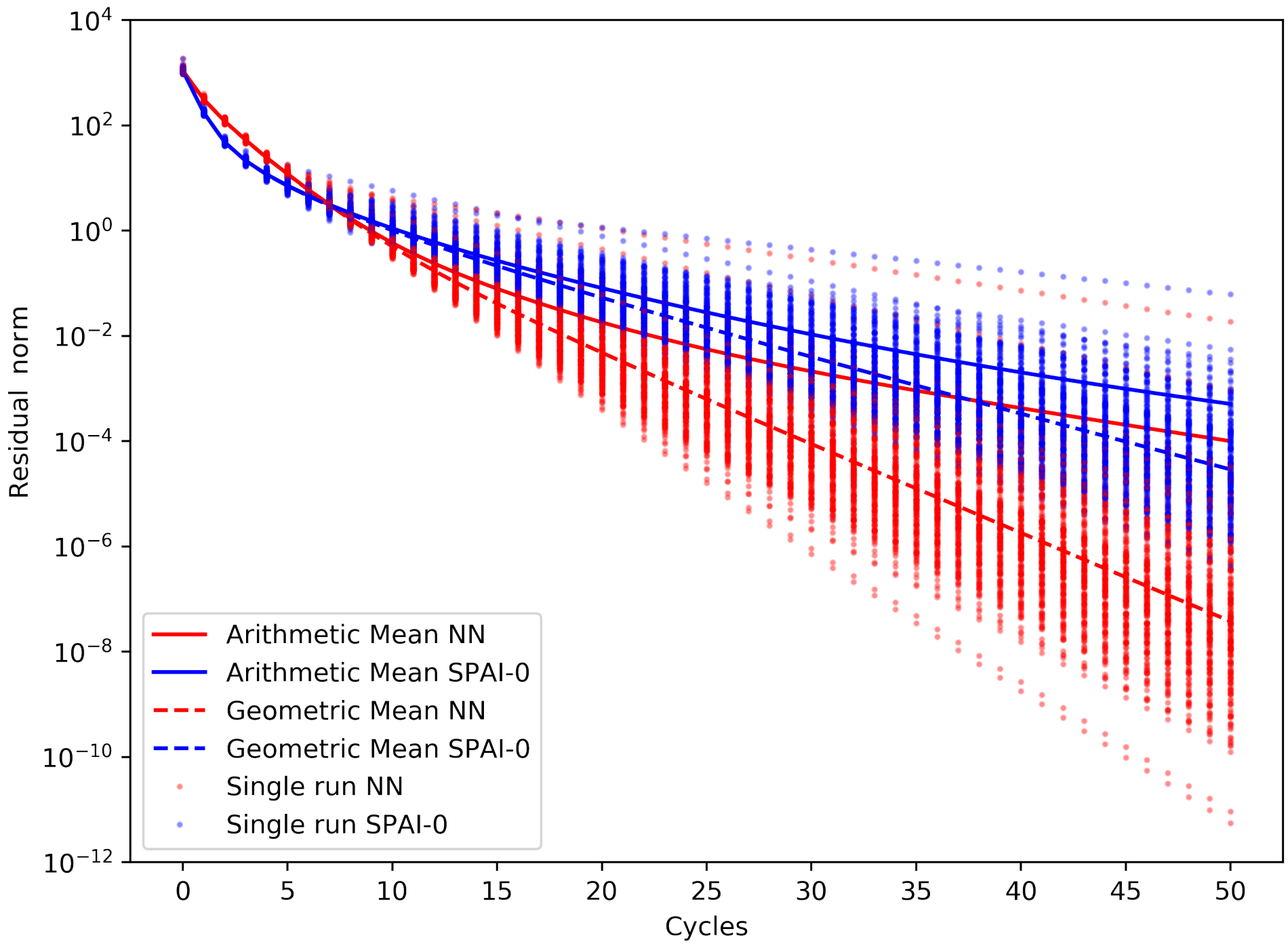

Figure 4.7: Results of SPAI-0 (blue) vs NN trained on normalized data (red). Averages were also computed: arithmetic averages (continuous lines) and geometric averages (dashed lines). The (y-axis) is the residual norm left after each cycle. The dots represent single data samples for NN and SPAI-0.

Table 4.6: Comparison of the convergence rates of SPAI-0 (last line) and the NN trained with different parameters on normalized data. *Pre* and *Post* are the numbers of relaxations performed respectively before and after the coarse grid correction. Columns *NN wins* show the percentage of samples for which NN outperforms SPAI-0. Columns *Conv. rate* show the geometrical average of convergence rates.

| Method | | | Results of testing | | | |
|---|---|---|---|---|---|---|
| Type | Training param. | | $\nu_1 = 1, \nu_2 = 0$ | | $\nu_1 = 2, \nu_2 = 1$ | |
| | $\nu$ | $\alpha$ | NN wins | Conv. rate | NN wins | Conv. rate |
| NN | 1 | 1 | 4.5% | 0.7800 | 18% | 0.6664 |
| NN | 1 | 2 | 99% | 0.7469 | 100% | 0.6438 |
| NN | 1 | 4 | 99% | 0.7153 | 100% | 0.6256 |
| NN | 1 | 6 | 99% | 0.6971 | 100% | 0.6159 |
| NN | 1 | 8 | 99% | 0.6788 | 100% | 0.6070 |
| NN | 1 | 10 | 99% | 0.6648 | 100% | 0.6017 |
| NN | 2 | 1 | 98% | 0.7591 | 99% | 0.6511 |
| NN | 2 | 2 | 98.5% | 0.7478 | 100% | 0.6444 |
| NN | 2 | 4 | 99.5% | 0.7449 | 100% | 0.6427 |
| NN | 2 | 6 | 99.5% | 0.7431 | 100% | 0.6414 |
| NN | 2 | 8 | 99.5% | 0.7411 | 100% | 0.6404 |
| NN | 2 | 10 | 99.5% | 0.7400 | 100% | 0.6399 |
| NN | 3 | 1 | 98% | 0.7565 | 99% | 0.6495 |
| NN | 3 | 2 | 99% | 0.7021 | 100% | 0.6185 |
| NN | 3 | 4 | 99% | 0.6424 | 100% | 0.5779 |
| NN | 3 | 6 | 93.5% | 0.6845 | 100% | 0.5613 |
| NN | 3 | 8 | 72% | 0.7085 | 100% | 0.5516 |
| NN | 3 | 10 | 62.5% | 0.7211 | 100% | 0.5438 |
| NN | 4 | 1 | 98.5% | 0.7537 | 99% | 0.6476 |
| NN | 4 | 2 | 99% | 0.7024 | 100% | 0.6186 |
| NN | 4 | 4 | 99% | 0.6851 | 100% | 0.6097 |
| NN | 4 | 6 | 99% | 0.6790 | 100% | 0.6071 |
| NN | 4 | 8 | 99% | 0.6754 | 100% | 0.6055 |
| NN | 4 | 10 | 99% | 0.6720 | 100% | 0.6042 |
| SPAI-0 | - | - | - | 0.7755 | - | 0.6647 |

### 4.2.3 NN as WJ

We saw in the last section that normalized NN outperform the SPAI-0 method. Our initial goal is reached, supposedly. But a closer look shows in the present section that the normalized NN did not actually use any information from the stencil. It functioned in fact as a finder of the optimal coefficient for WJ. The present section consists in three parts: (i) normalized NN vs. Weighted Jacobi, (ii) an alternative view of the normalized NN, and (iii) evidence that the normalized NN is equivalent to the Weighted Jacobi coefficient finder.

The work here starts from a comparison of normalized NN-based relaxation with Weighted Jacobi. The intention is to reach better Multigrid convergence rates with NN-based relaxation. Despite not reaching that goal, an important insight is gained for the continuation of the work. In this part, the best[2] normalized NN from section 4.2.2 is compared with the best result achieved by the Weighted Jacobi method. Brute force is employed to find the coefficient $\omega$ (as defined in section 2.1) that optimizes the convergence rate of Weighted Jacobi. For the similar testing parameters as in the previous part (see table 4.5) excluding prolongation method that changed to Blackbox, the results for the normalized NN and for the Weighted Jacobi method are found respectively in tables 4.7 and 4.8 for $\nu_1 = 2, \nu_2 = 1$ relaxation. As can be clearly seen from these tables, the NN reaches almost the same convergence rate values as Weighted Jacobi.

A deeper look at the normalized NN shows a possible explanation for the close values of convergence rates,

$$m_{k_{norm}} = \underbrace{\frac{1}{a_{kk}}}_{D^{-1}} \cdot \underbrace{f'_{NN}(a_{k1}/a_{kk}, \ldots, a_{kn}/a_{kk})}_{const\ ?} \overset{?}{\Longleftrightarrow} M_{WJ} = \omega \cdot D^{-1}. \tag{4.5}$$

This equation shows that when $f'_{NN}$ returns a constant value, it has the exact same structure as Weighted Jacobi.The confirmation consists of observing that normalization involves two changes: the first is the multiplication of the outcome of the NN by $1/a_{kk}$ and the second is the division of all values of the stencil input to the NN by the value on the diagonal $a_{kk}$. The normalization is decomposed as follows: multiplication by $1/a_{kk}$ (case 1, Eq. (4.6)) and division of the inputs by $a_{kk}$ (case 2, Eq. (4.7)) in order to examine the corresponding contributions to the convergence of the algorithm,

$$m_{k_{case1}} = \frac{1}{a_{kk}} \cdot f_{NN}(a_{k1}, \ldots, a_{kn}), \tag{4.6}$$

$$m_{k_{case2}} = f'_{NN}(a_{k1}/a_{kk}, \ldots, a_{kn}/a_{kk}). \tag{4.7}$$

Additional training was performed with the same parameters as in table 4.4. Here the

[2] i.e., the one which achieves the lowest convergence rate value.

Table 4.7: Comparison of the convergence rates of NN (first line, one with the best convergence rate among all pre-trained NNs) and the WJ with the optimal coefficient value found by brute force. The numbers of relaxations performed are 1 before and 0 after the coarse grid correction. Column *Conv. rate* show the geometrical average of convergence rates.

| Type | Parameters | | Conv. rate |
|---|---|---|---|
| NN | $\nu = 1$ | $\alpha = 25$ | 0.6304 |
| WJ | $\omega = 1.00$ | | 0.6972 |
| WJ | ... | | ... |
| WJ | $\omega = 1.07$ | | 0.6411 |
| WJ | $\omega = 1.08$ | | 0.6335 |
| WJ | $\omega = 1.09$ | | 0.6305 |
| WJ | $\omega = 1.10$ | | 0.6322 |
| WJ | $\omega = 1.11$ | | 0.6388 |

Table 4.8: Same as Table 4.7 but with $\nu_1 = 2, \nu_2 = 1$ relaxations.

| Type | Parameters | | Conv. rate |
|---|---|---|---|
| NN | $\nu = 3$ | $\alpha = 25$ | 0.5259 |
| WJ | $\omega = 1.00$ | | 0.6178 |
| WJ | ... | | ... |
| WJ | $\omega = 1.18$ | | 0.5426 |
| WJ | $\omega = 1.19$ | | 0.5307 |
| WJ | $\omega = 1.20$ | | 0.5219 |
| WJ | $\omega = 1.21$ | | 0.5237 |
| WJ | $\omega = 1.22$ | | 0.5361 |

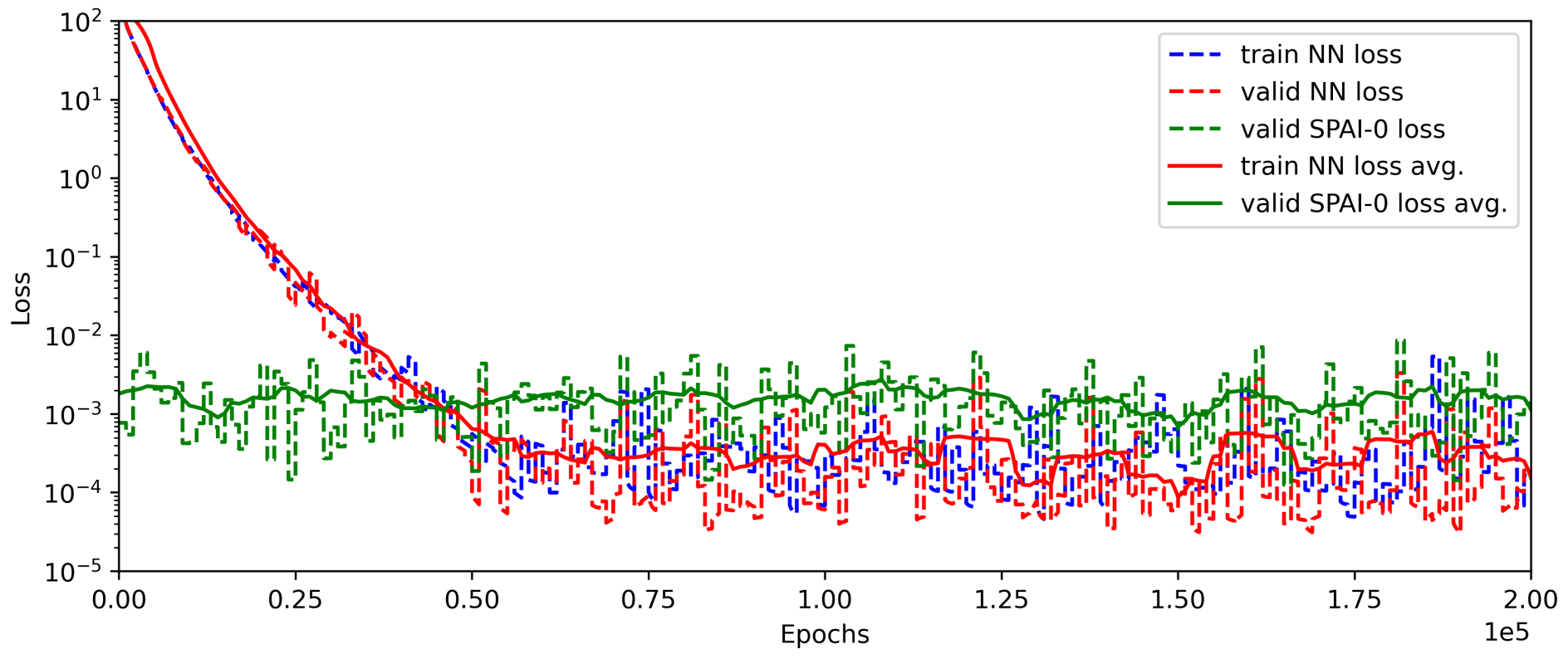

(a)

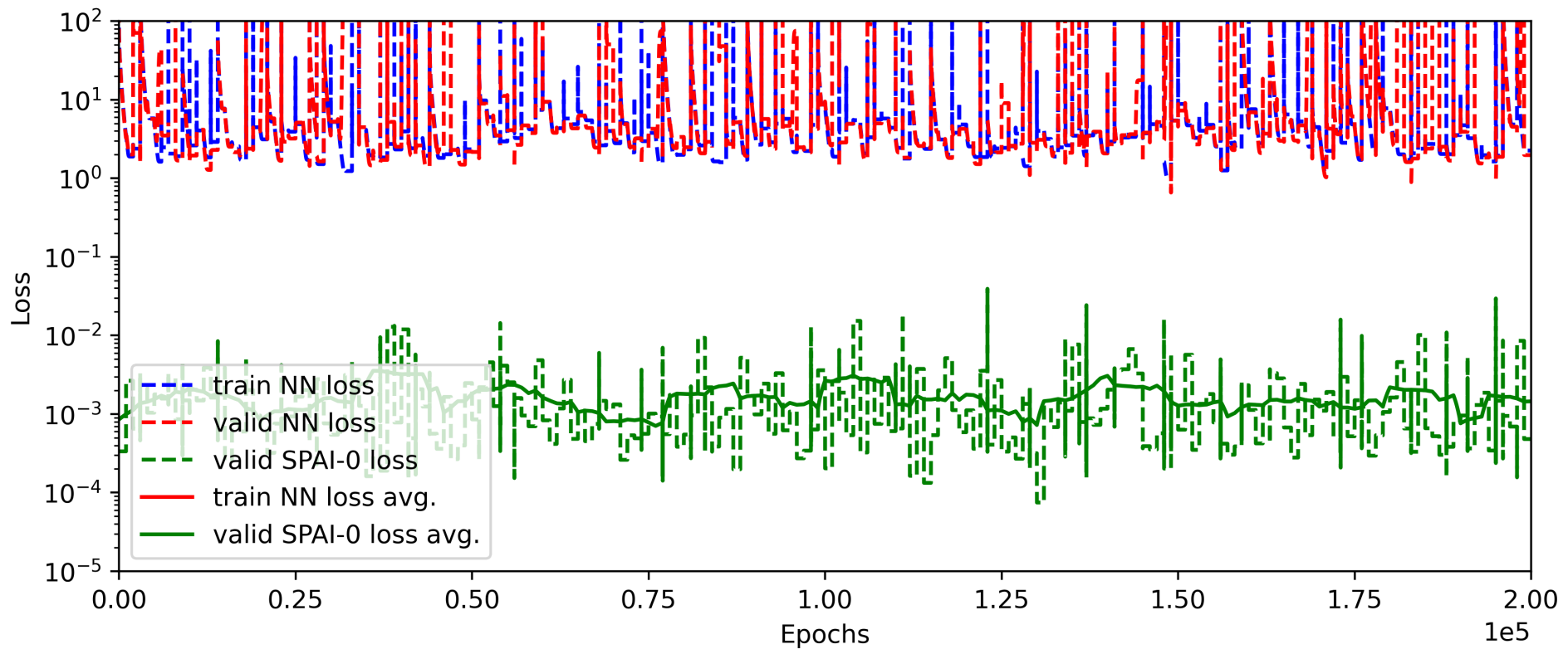

(b)

Figure 4.8: Plots (a) and (b) represent the loss functions corresponding to the training and validation of the NN (respectively blue and red dashed lines), and to the validation implemented with SPAI-0 (green dashed line). The (x-axis) represents the number of epochs. The (y-axes) plot the training losses. Figures (a) and (b) refer to Case 1 and Case 2 respectively. The same validation data sets were used for both NN and SPAI-0. The $1e4$ average values are represented by solid curves in the same colors.

training process of section 4.2.2 is run once with case 1 and again with case 2. The results speak for themselves as can be seen from Fig. 4.8. With case 1 (4.8a), training the NN with $m_{k_{case1}}$ led to results as good as in the normalized case, whereas with case 2 (4.8b), $m_{k_{case2}}$ led to outcomes worse than those of the non-normalized algorithm.

In order to verify the hypothesis (Normalized NN from section 4.2.2 is mostly a WJ optimiser) we build a new narrow NN and analyze it during the training and testing steps. This NN has a single neuron with a single weight whose input is a constant, in other words this NN has no input. At the training stage, the narrow NN behaves as well as the regular NN and even reaches lower loss than SPAI-0 with much less cycles, Figure 4.9. Test results of this NN are similar as well. For example the convergence rate of narrow NN after 47000 to 50000 cycles are 0.6594 to 0.6305 (table 4.9), while for Normalized NN it is 0.6305. The convergence rate value jumps of narrow NN can be explained by fact that the whole narrow NN is virtually a single neuron, which changes relatively quickly its value depending on the current batch of data. This issue is relatively easy to solve by additional adaptation of the learning rate dependent on training progresses. This was not done because the aim of the experiment had been served.

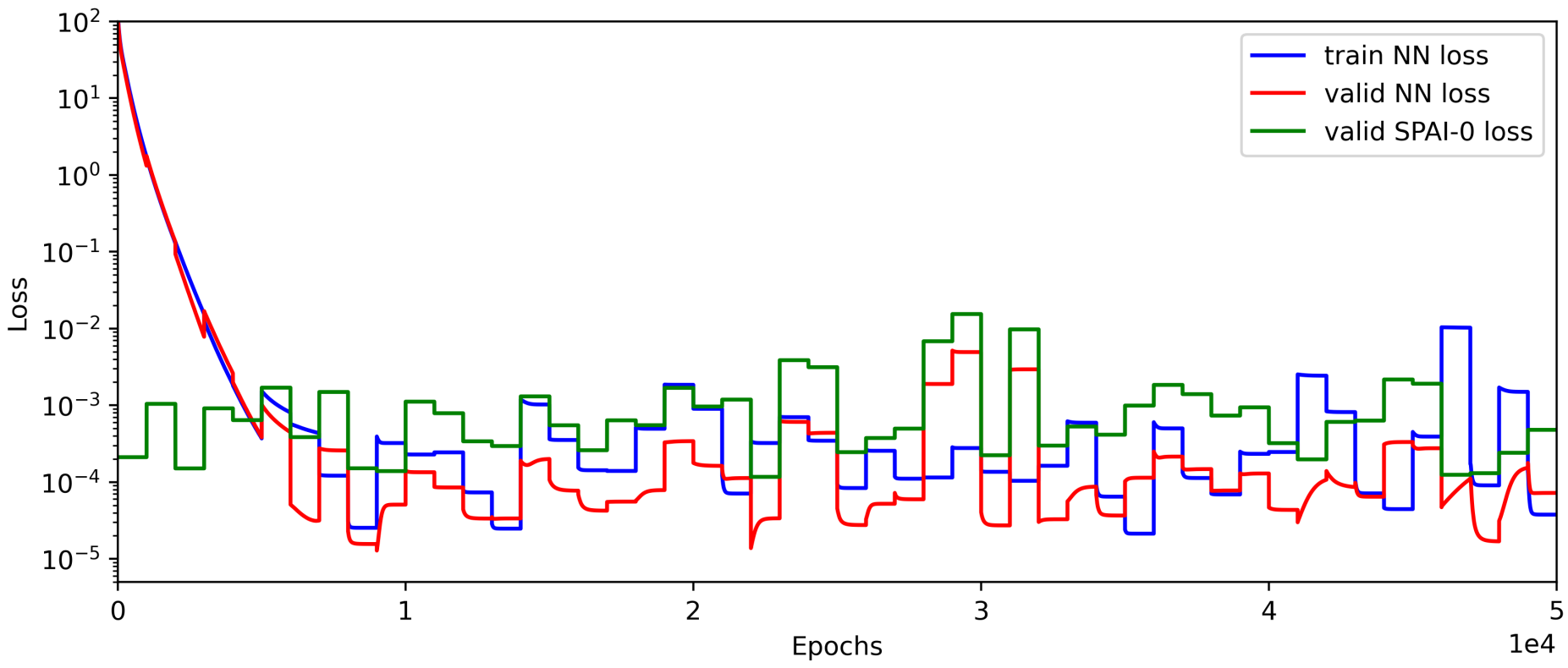

Figure 4.9: This plot shows the loss functions corresponding to the training and validation of the narrow NN (respectively blue and red lines), and to the validation implemented with SPAI-0 (green line). The (x-axis) represents the number of epochs. The (y-axes) plot the training losses. The same validation data sets were used for both NN and SPAI-0.

Table 4.9: Comparison of the convergence rates of normalised NN (first line) and the narrow NN after different amounts of epochs. Both of the NNs were trained with the same parameters ($\alpha = 1$, $\nu = 25$). The numbers of relaxations performed are 1 before and 0 after the coarse grid correction. Column *Conv. rate* shows the geometrical average of convergence rates.

| Type | Epochs | Conv. rate |
|---|---|---|
| Norm. NN | $2.00e5$ | 0.6304 |
| Narrow NN | $4.70e4$ | 0.6362 |
| Narrow NN | $4.75e4$ | 0.6305 |
| Narrow NN | $4.80e4$ | 0.6314 |
| Narrow NN | $4.85e4$ | 0.6443 |
| Narrow NN | $4.90e4$ | 0.6564 |
| Narrow NN | $4.95e4$ | 0.6584 |
| Narrow NN | $5.00e4$ | 0.6594 |

As a conclusion, to get results as good as those of the NN, it is enough to use a WJ algorithm with an optimal value of $\omega$. The coefficient $\omega$ can be found using one of the well known optimization techniques such as local search, random search or even brute force. Our research shows that the "information" of 9 stencil values that we thought would be used by the NN function $f_{nn}$ to improve the relaxation is basically not being used. All that the NN does is to find a constant result for $f_{nn}$. This constant is dependent on the setup of the problem and on various factors like the number of relaxations, the type of prolongation, the data distribution, etc. In order to minimize the spectral radius, a WJ algorithm which uses only the diagonal value, $a_{kk}$ of $A$ in order to calculate $M$ is enough. In others words, there are redundancies in the "information" contained in the 9-value stencil. The comparison of our results for the Weighted Jacobi relaxation method show that WJ got similar results as our method based on a NN. Moreover, WJ has a much simpler calculation for the same outcome.

We expand the validity of this result in the following section in which anisotropic grids are introduced.

### 4.2.4 Anisotropic Scenario

This section examines whether the stencil-based NN can distinguish locally between isotropic and anisotropic grids and thereby improve the overall convergence rate for different grid aspect ratios. Isotropic grids are a special case of anisotropic grids with $h_x = h_y$ as explained in detail in section 2.2.

This study is similar to the one executed in section 4.2.2. In order to focus the investigation, all other parameters are kept unchanged. A basic configuration is investigated in which just two aspect ratios are used. It consists of isotropic samples ($h_x = h_y = 1$, as have been used in all the previous experiments) and anisotropic samples with $h_x = 1, h_y = 2$.

The training setup is similar to that of section 4.2.2. The main difference is that the NN is trained on a mix of isotropic and anisotropic samples. Each batch of training data includes isotropic and anisotropic samples in a normal distribution proportion, whereby 50% of the samples of each dataset are isotropic on average while the rest are anisotropic. The training phase performed properly without any remarkable event.

As in section 4.2.2, the convergence rate of the NN-based multigrid relaxation is compared to that of WJ (using exhaustive search for the optimal $\omega$). The test includes three runs: an isotropic dataset, an anisotropic dataset, and a mixed dataset (normally

Table 4.10: Convergence rates of multigrid for NN and WJ relaxation components when isotropic, anisotropic and a mix of isotropic and anisotropic data samples are used (testing phase). The NN was trained on a mix of isotropic and anisotropic data samples. $\omega$ is the DJ coefficient.

| type | | Convergence rate | | |
|---|---|---|---|---|
| | $\omega$ | isotropic | anisotropic | mixed |
| WJ | 0.71 | 0.6431 | 0.7539 | 0.6977 |
| WJ | 0.72 | 0.6384 | 0.7313 | 0.6844 |
| WJ | 0.73 | 0.6336 | 0.7386 | 0.6851 |
| WJ | 0.74 | 0.6288 | 0.7602 | 0.6928 |
| WJ | ... | ... | ... | ... |
| WJ | 0.96 | 0.5056 | 1.2780 | 0.8175 |
| WJ | 0.97 | 0.4917 | 1.3016 | 0.8097 |
| WJ | 0.98 | 0.4845 | 1.3251 | 0.8119 |
| WJ | 0.99 | 0.4868 | 1.3487 | 0.8236 |
| NN | — | 0.6376 | 0.7304 | 0.6835 |

distributed as in the training phase).

Table 4.10 shows that the value of $\omega$ that optimizes the convergence rate for WJ depends significantly on the aspect ratio. This means that if the NN is able to discriminate between anisotropic and isotropic samples, it can return an optimal coefficient for that specific sample and thus improve the overall convergence rate. From the same table, it appears that the stencil-based NN (last line) does not show any ability to discriminate between the isotropic and anisotropic cases. As a matter of fact, the NN reaches convergence rates that are no better than those of the WJ algorithm with the optimal coefficient for the mix case, $\omega = 0.72$.

The general picture of the results obtained with stencil-based NNs and the results of section 4.2.3 in particular encourage us to reconsider the use of neural networks for relaxation schemes with a diagonal preconditioner, because the additional information that the 9-point stencil provides does not provide an added value. It is possible that with a different data distribution (e.g., real life data) the stencil-based NN may lead to better results. A bigger NN might also bring about improvements (for example inputs of the NN could include neighboring stencils as well) but this direction of research seems impractical due to its high computational cost.

## 4.3 NN for 4-Color relaxation

As seen in the previous section, finding the optimal coefficient $\omega$ of WJ may improve the convergence rates significantly, as shown in tables 4.7 and 4.8 for example. However, the optimal $\omega$ depends on various factors, such as the multigrid configuration (e.g., the number of relaxations or the prolongation type) and the data distribution. The optimal $\omega$ can be found, relatively easily, using one of many optimization techniques. However, it may be possible to obtain far better convergence rates with comparable run-time by using a more elaborate relaxation scheme, at the cost of optimizing several parameters rather than just one. We study this option in the present subsection.

The 4-Color Gauss–Seidel relaxation, described in detail in section 2.3, is a likely candidate for our problem. The question we wish to address here is whether relaxation parameters computed by a NN can improve convergence rates significantly. The most straightforward approach is to use a single parameter $\omega$ for all four colors, which is essentially as simple as finding the optimal $\omega$ for Jacobi relaxation. A more ambitious approach for 4-Color SOR is presented in Alg. 4.1. Here, we assign a different value $\omega_i$ to each color, thus generalizing the single-value configuration of Alg. 2.2. This increases the complexity of optimizing the parameters, but, as we show below, results in faster convergence. Since the optimization is performed once and for all (for a given distribution) and the resulting parameters may be reused indefinitely, this effort pays off. Optimization of these coefficients by brute force is impractical. An alternative, which

**Algorithm 4.1** Proposed 4-Color Successive over-relaxation

**Input:**

i Discretization of matrix $A$.

ii Current solution approximation $u^{(k)}$

iii Right-hand side $f$

iv Relaxation coefficients $[\omega_1,\omega_2,\omega_3,\omega_4]$

**Output:**

i Result, approximation after relaxation $u^{(k+1)}$

**Steps:**

1: Perform Weighted Jacobi relaxation of all “1” color nodes,
$u_i^{(k)} := (1-\omega_1)u_i^{(k)} + \omega_1 \cdot \frac{1}{a_{ii}}(f_i - \sum_{j\in\{n_i\}} a_{i,j}u_j^{(k)})$

2: Perform Weighted Jacobi relaxation of all “2” color nodes,
$u_i^{(k)} := (1-\omega_2)u_i^{(k)} + \omega_2 \cdot \frac{1}{a_{ii}}(f_i - \sum_{j\in\{n_i\}} a_{i,j}u_j^{(k)})$

3: Perform Weighted Jacobi relaxation of all “3” color nodes,
$u_i^{(k)} := (1-\omega_3)u_i^{(k)} + \omega_3 \cdot \frac{1}{a_{ii}}(f_i - \sum_{j\in\{n_i\}} a_{i,j}u_j^{(k)})$

4: Perform Weighted Jacobi relaxation of all “4” color nodes,
$u_i^{(k)} := (1-\omega_4)u_i^{(k)} + \omega_4 \cdot \frac{1}{a_{ii}}(f_i - \sum_{j\in\{n_i\}} a_{i,j}u_j^{(k)})$

5: Return $u^{(k+1)} := u^{(k)}$

$\{n_i$ refer to neighboring nodes of node $i\}$

we adopt, is to use the same technique as in section 4.2, namely neural networks, to provide not one but a set of 4 coefficients. The contribution proposed in the present the work is a NN-based tool for obtaining optimized coefficients for 4-Color SOR employed in multigrid solvers.

The proposed algorithm consists of a training phase using a narrow NN similar to that of section 4.2.3 but with four coefficients, followed by the application of a few iterations of fine tuning with a local search genetic algorithm. (In practice always found that the narrow NN returned coefficients in the vicinity of the global optimum). This second step yields significant improvement.

Table 4.11: Training parameters

| Parameters | |
|---|---|
| Method | Two-grid |
| grid size | 16 × 16 |
| data distribution | log normal |
| number of samples | 100 |
| train / test division | 0.8 / 0.2 |
| normalization by row | true |
| loss function | mean square |
| $k$ | 40 |
| $\delta$ | 0.0001 |
| $\nu$ | 1 |
| Prolongation | Black-Box |
| Relaxation order | 4-Color |
| Relaxation type | SOR |

The training of the NN is similar to that of section 4.2.3. The training parameters are provided in table 4.11. The input to the NN are the "color number", and the output is the corresponding coefficients for each color. In order to assess the stability of the training process as a function of grid size, training on $8 \times 8$, $32 \times 32$ and $64 \times 64$ grid is performed. To answer the question of whether there are additional coefficients for which a better minimum can be reached, the NN is trained with different random initialization values. This decreases the likelihood that we missed a better minimum and thus improves robustness.

Table 4.12: Coefficients of 4-Color SOR for training on different grid sizes

| grid size | Coef. 1 | Coef. 2 | Coef. 3 | Coef. 4 |
|---|---|---|---|---|
| 8 × 8 | 0.6882 | 1.0807 | 1.0702 | 1.0308 |
| 16 × 16 | 0.7319 | 1.0968 | 1.0955 | 1.0444 |
| 32 × 32 | 0.7539 | 1.1145 | 1.1141 | 1.0513 |
| 64 × 64 | 0.7558 | 1.1191 | 1.1188 | 1.0525 |

Table 4.12 shows that coefficient values converge and that training on $32 \times 32$ grids is sufficient in practice in order to find the coefficients near the optimum. The reason why we can train on small grids and obtain coefficients that are useful for much finer grids is that the error-smoothing process is local. Therefore, the relaxation can be optimized locally, provided that the grid used in training are not so small such that boundary effects are significant. All the runs executed with different random initialization values (16 runs in total) converge to the same values (fig. 4.11), indicating the robustness of the optimization process.

Testing was executed with the parameters of table 4.13. This step does not require any NN, just implementing Algorithm 4.1 in the relaxation phase with the optimized relaxation parameters. That is, the testing does not require any further calculation once the optimal coefficients for the prescribed distribution of diffusion coefficients have been established. In addition to a "W" multigrid cycle, testing was also conducted with "V" and "F" cycles. Testing was also done on 128 × 128 and 256 × 256 grids in order to assess performance of the proposed method on larger grids.

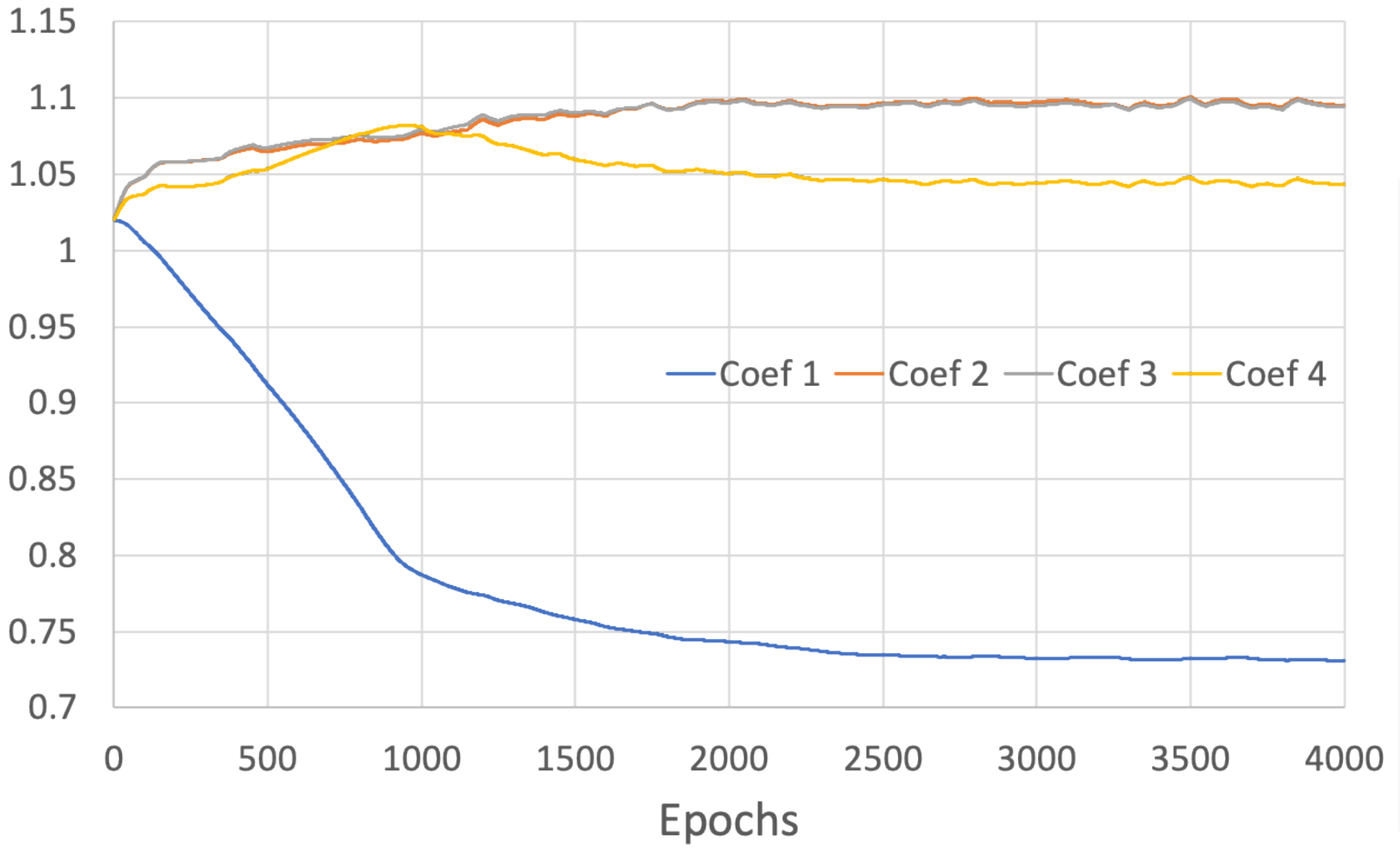

Figure 4.10: Coefficients of 4-Color SOR for training as a function of epochs. The initial value is the same for all the coefficients.

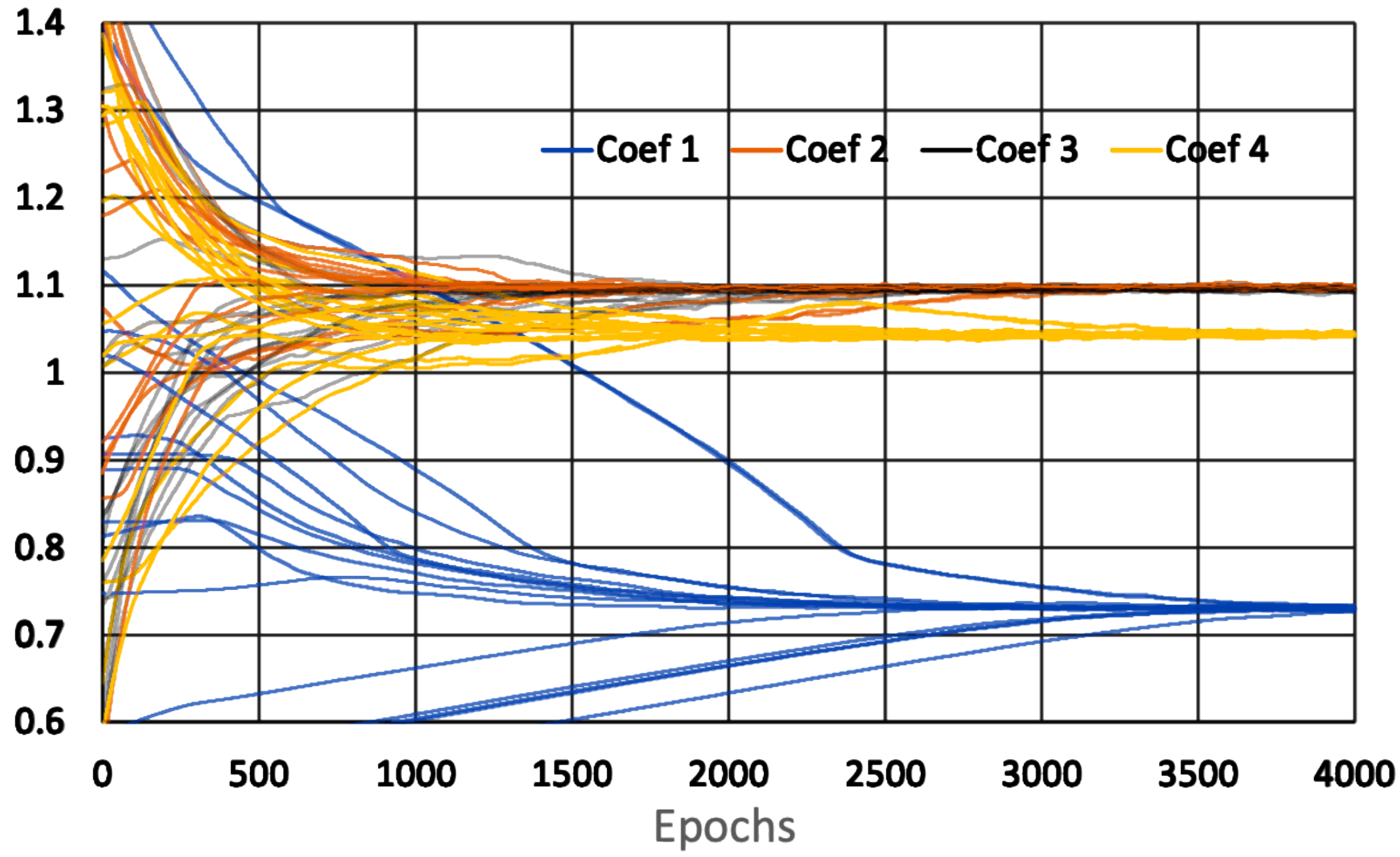

Figure 4.11: Coefficients of 4-Color SOR as a function of epochs for 16 training runs conducted using random initial values. This experiment indicates the robustness of the NN in providing consistent coefficients for 4-Color SOR.

Table 4.13: Testing parameters

| Parameters | |
|---|---|
| Method | Multigrid |
| grid size | 64 × 64 |
| data distribution | log normal |
| number of samples | 10 |
| $\delta$ | 0 |
| Pre-Relaxation | 1 |
| Post-Relaxation | 0 |
| Prolongation | Black-Box |
| Relaxation order | 4-Color |
| Relaxation type | SOR |
| Cycle type | W |

At the test stage the mean convergence rate per cycle is about 1/7, with only a single relaxation per level ($\nu = 1$), for a log-normal distribution of the random diffusivity coefficients. Compare to the simple Poisson problem with the standard 5-point star discretization, where the best known method is Red-Black relaxation which achieves a convergence rate of about 1/4 for $\nu = 1$. We find that the performance achieved by the NN for our far more challenging problem is remarkably good. This results are compared to the results of Alg. 2.2 with $\omega = 1$ and with a single common $\omega$ optimized by brute force. The results for a "W" cycle with grid size 64 × 64 are presented in table 4.14. The proposed algorithm (with training on 64 × 64 grids) shows improvements of more than 16% and 38% compared to optimal single coefficient SOR and to Gauss-Seidel ($\omega = 1$), respectively. A local search on these results ($NN_{64\times64} \pm 0.01$) improves the results by more than one percent, just shy of the minimum found using a local search genetic algorithm ("brute force"). To validate these results, three additional tests on other datasets with the same configurations were performed (appendix A.6). The "F" cycle test reaches results that are similar to those of the "W" cycle, with improvements of about 10% compared to the optimal "Common" case and 33% compared to Gauss-Seidel ($\omega_i = 1$). The results with "V" cycles are not as good. Results of tests with the "F" and "V" cycles are available in appendices A.7 and A.8, respectively. Tests on a bigger grid size (256 × 256) show a similar picture, see in table 4.15. This proves once again that the values that were found during the training phase remain valid for larger grids.

The proposed method is also applied to the Poisson equation mentioned in section

Table 4.14: W-cycle for testing on grid size $64 \times 64$. *Common* refers to 4-Color SOR with identical input parameters. $\text{NN}_{32\times32}$ and $\text{NN}_{64\times64}$ indicate the size of the grid used for training, respectively $32 \times 32$ and $64 \times 64$. $\text{NN}_{64\times64} \pm 0.01$ indicates best values after conducting a local search algorithm for training. *Brute force* indicates best values found using a genetic search algorithm for training. The *Improvement* columns indicate the number of testing cycles required to reach the same residual as a proportion of the number of cycles required by two common 4-Color SOR processes (indicated in the columns by 100%).

| Type | 4-Colors values | | | | Convergence Rate | Improvement | |
|---|---|---|---|---|---|---|---|
| Common | 1.00 | 1.00 | 1.00 | 1.00 | 0.3044 | — | 100% |
| Common | ··· | ··· | ··· | ··· | — | — | — |
| Common | 1.070 | 1.070 | 1.070 | 1.070 | 0.2053 | — | 75.13% |
| Common | 1.080 | 1.080 | 1.080 | 1.080 | 0.1986 | 100% | 73.58% |
| Common | 1.090 | 1.090 | 1.090 | 1.090 | 0.2037 | — | 74.77% |
| $\text{NN}_{32\times32}$ - exact | 0.754 | 1.114 | 1.114 | 1.051 | 0.1477 | 84.52% | 62.19% |
| $\text{NN}_{64\times64}$ - exact | 0.756 | 1.119 | 1.119 | 1.052 | 0.1438 | 83.37% | 61.34% |
| $\text{NN}_{64\times64} \pm 0.01$ | 0.766 | 1.129 | 1.129 | 1.052 | 0.1386 | 81.82% | 60.20% |
| Brute force | 0.766 | 1.139 | 1.129 | 1.052 | 0.1340 | 80.44% | 59.18% |

Table 4.15: W-cycle for testing on grid size $256 \times 256$. Same structure as table 4.14. Additional local search algorithms for training were conducted.

| Type | 4-Colors values | | | | Convergence Rate | Improvement | |
|---|---|---|---|---|---|---|---|
| Common | 1.00 | 1.00 | 1.00 | 1.00 | 0.3543 | — | 100% |
| Common | ··· | ··· | ··· | ··· | — | — | — |
| Common | 1.090 | 1.090 | 1.090 | 1.090 | 0.2242 | — | 69.39% |
| Common | 1.100 | 1.100 | 1.100 | 1.100 | 0.2140 | 100% | 67.31% |
| Common | 1.110 | 1.110 | 1.110 | 1.110 | 0.2212 | — | 68.78% |
| $\text{NN}_{64\times64}$ - exact | 0.756 | 1.119 | 1.119 | 1.052 | 0.2038 | 96.92% | 65.24% |
| $\text{NN}_{64\times64} \pm 0.01$ | 0.766 | 1.129 | 1.129 | 1.062 | 0.1788 | 89.56% | 60.28% |
| $\text{NN}_{64\times64} \pm 0.02$ | 0.776 | 1.139 | 1.139 | 1.072 | 0.1639 | 85.25% | 57.38% |
| $\text{NN}_{64\times64} \pm 0.03$ | 0.786 | 1.149 | 1.149 | 1.072 | 0.1545 | 82.54% | 55.56% |
| Brute force | 0.776 | 1.159 | 1.159 | 1.062 | 0.1473 | 80.49% | 54.18% |

2.2. The entire process of training a narrow 4-color NN on a $64 \times 64$ grid and testing the optimal coefficients that are found is repeated. The optimal coefficients for Poisson equation found by the NN are $0.6404, 1.0632, 1.0086$ and $0.9831$ for coefficients 1 to 4 respectively.

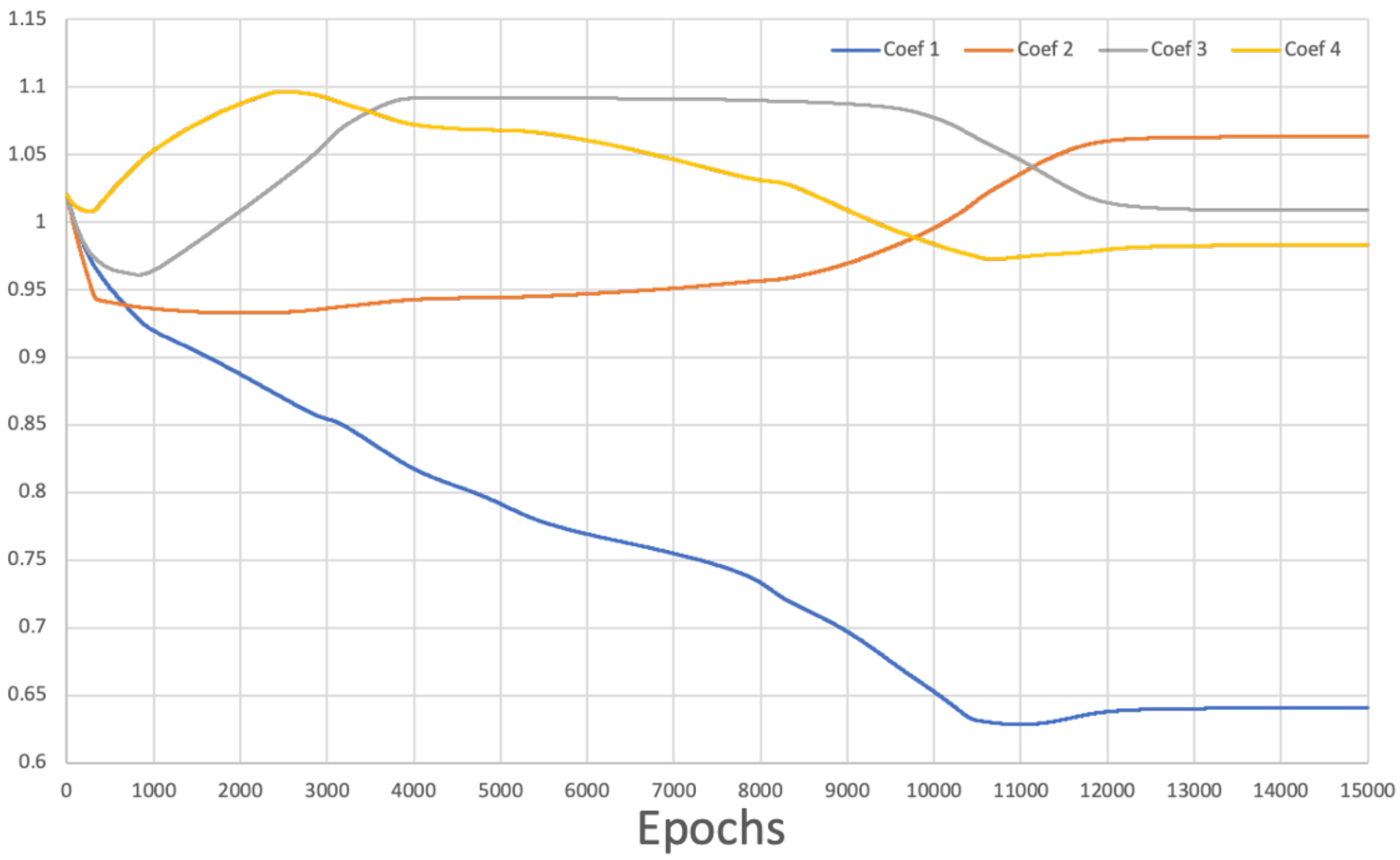

Figure 4.12: Coefficients of 4-Color SOR for training as a function of epochs to find the solution to the Poisson equation, a special case of diffusion equation. The initial value is the same for all the coefficients.

Table 4.16: F-cycle for testing the Poisson equation on grid size $64 \times 64$. The structure is the same as table 4.14

. .

| Type | 4-Colors values | | | | Convergence Rate | Improvement | |
|---|---|---|---|---|---|---|---|
| Common | 1.00 | 1.00 | 1.00 | 1.00 | 0.1302 | — | 100% |
| Common | ... | ... | ... | ... | — | — | — |
| Common | 0.980 | 0.980 | 0.980 | 0.980 | 0.1197 | — | 96.01% |
| Common | 0.970 | 0.970 | 0.970 | 0.970 | 0.1159 | 100% | 94.58% |
| Common | 0.960 | 0.960 | 0.960 | 0.960 | 0.1222 | — | 96.96% |
| $NN_{64\times64}$ - exact | 0.640 | 1.063 | 1.009 | 0.983 | 0.0727 | 82.22% | 77.77% |
| $NN_{64\times64} \pm 0.01$ | 0.640 | 1.053 | 1.019 | 0.973 | 0.0723 | 82.02% | 77.57% |
| Brute force | 0.630 | 1.053 | 1.029 | 0.973 | 0.0719 | 81.89% | 77.45% |

Table 4.16 shows that the gains relatively to the “Common” case are comparable, which suggests the universality of the proposed method. This will be reviewed and summarized in the following chapter.

# Chapter 5

# Conclusion

In this thesis we report our study on the application of neural networks for optimizing relaxation in multigrid algorithms for diffusion problems with random coefficients. For log-normally distributed diffusion coefficients in 2D, we achieve an asymptotic convergence rate of about 1/7 with just a single relaxation per level. This exceptional convergence rate is achieved by 4-color SOR with four different relaxation parameters optimized simultaneously using a NN.

In the first part of the report we describe attempts at using a stencil-based NN, intended to serve as a generator of sparse inverse matrices for multigrid relaxation. Although this step did not bring about the expected improvements, important conclusions used in the sequel were drawn from it. The second part describes narrow NNs used for optimizing coefficients for weighted Jacobi relaxation (WJ) and 4-color SOR. The main result is achieved by optimizing multiple SOR coefficients for 4-Color relaxation. This approach yields significant improvements in the convergence rate of multigrid algorithms, at negligible additional run-time cost once the coefficients have been precomputed once and for all. Specifically, average convergence rates are improved by up to 40% compared to standard (unweighted) 4-color Gauss-Seidel relaxation and by almost 20% over the best 4-Color mono-value relaxation.

Possible directions of future research include: algorithms that optimize 4-Color SOR coefficients for each multigrid layer separately; separation of coefficients by cycles, for example, into even and odd cycles to find unique sets of optimized coefficients to each of the sets; integration of the method presented here with NN based prolongation, cf., [GGB+19]; extension to nonsymmetric and indefinite problems, implementation on unstructured grids and learning relaxations for systems of partial differential equations.

# Appendix A

# Appendix

## A.1 Condition of convergence for Jacobi and Gauss-Seidel iterative methods

If the matrix $A$ is strictly or irreducibly diagonally dominant, then the Jacobi and Gauss-Seidel methods are guaranteed to converge, [Bag95]. This is a sufficient condition, not a necessary one. In particular, the convergence of Gauss-Seidel is guaranteed for any $A$ that is symmetric positive-definite, [GVL96] Theorem 10.1.2.

## A.2 Black Box method

The creation of prolongation matrix $P$ in the case of Bilinear prolongation is simple because it is built with constant values, cf., Fig 4.1.

For the restricted point (coarse finite element) in the case of the Bilinear restriction ($P^T$), with $c_{j,i}$ and $a_{j,i}$, semi-centered points of coarse and fine grids respectively,

$$\begin{aligned} c_{j,i} = P_c \cdot a_{j,i} + P_n \cdot a_{j+\frac{1}{2},i} + P_s \cdot a_{j-\frac{1}{2s},i} + P_e \cdot a_{j,i+\frac{1}{2}} + P_s \cdot a_{j,i-\frac{1}{2}} \\ +P_{ne} \cdot a_{j+\frac{1}{2},i+\frac{1}{2}} + P_{nw} \cdot a_{j+\frac{1}{2},i-\frac{1}{2}} + P_{se} \cdot a_{j-\frac{1}{2},i+\frac{1}{2}} + P_{sw} \cdot a_{j-\frac{1}{2},i-\frac{1}{2}} \end{aligned} \tag{A.1}$$

where

$$\begin{aligned} P_c = \frac{1}{4} \\ P_n = P_s = P_e = P_w = \frac{1}{8} \\ P_{ne} = P_{nw} = P_{se} = P_{sw} = \frac{1}{16}. \end{aligned} \tag{A.2}$$

By comparison, in the case of Black Box, the P values vary for each point of the

grid:

$$
\begin{aligned}
P_c &= 1 \\
P_n &= \frac{a^n_{j-\frac{1}{2},i-\frac{1}{2}} + a^n_{j-\frac{1}{2},i} + a^n_{j-\frac{1}{2},i+\frac{1}{2}}}{a^n_{j,i-\frac{1}{2}} + a^n_{j,i} + a^n_{j,i+\frac{1}{2}}} \\
P_e &= \frac{a^e_{j+\frac{1}{2},i-\frac{1}{2}} + a^e_{j,i-\frac{1}{2}} + a^e_{j-\frac{1}{2},i-\frac{1}{2}}}{a^e_{j+\frac{1}{2},i} + a^e_{j,i} + a^e_{j-\frac{1}{2},i}} \\
P_s &= \frac{a^s_{j+\frac{1}{2},i-\frac{1}{2}} + a^s_{j+\frac{1}{2},i} + a^s_{j+\frac{1}{2},i+\frac{1}{2}}}{a^s_{j,i-\frac{1}{2}} + a^s_{j,i} + a^s_{j,i+\frac{1}{2}}} \\
P_w &= \frac{a^w_{j+\frac{1}{2},i+\frac{1}{2}} + a^w_{j,i+\frac{1}{2}} + a^w_{j-\frac{1}{2},i+\frac{1}{2}}}{a^w_{j+\frac{1}{2},i} + a^w_{j,i} + a^w_{j-\frac{1}{2},i}}.
\end{aligned} \tag{A.3}
$$

where $n$ in $a^n$ denotes the neighbor to the North of the current point of calculation and its own stencil. $e$, $s$, $w$, $ne$, $nw$, $se$ and $sw$ indicate the neighbors to the East, South, West, NorthEast, NorthWest, SouthEast and SouthWest of the current point of calculation, respectively.

$$
\begin{aligned}
P_{ne} &= -\frac{a^{ne}_{j-\frac{1}{2},i-\frac{1}{2}} + P_n \cdot a^{ne}_{j,i-\frac{1}{2}} + P_e \cdot a^{ne}_{j-\frac{1}{2},i}}{a^{ne}_c} \\
P_{nw} &= -\frac{a^{nw}_{j-\frac{1}{2},i+\frac{1}{2}} + P_w \cdot a^{nw}_{j-\frac{1}{2},i} + P_n \cdot a^{nw}_{j,i+\frac{1}{2}}}{a^{nw}_c} \\
P_{se} &= -\frac{a^{se}_{j+\frac{1}{2},i-\frac{1}{2}} + P_e \cdot a^{se}_{j+\frac{1}{2},i} + P_s \cdot a^{se}_{j,i-\frac{1}{2}}}{a^{se}_c} \\
P_{sw} &= -\frac{a^{sw}_{j+\frac{1}{2},i+\frac{1}{2}} + P_s \cdot a^{sw}_{j,i+\frac{1}{2}} + P_w \cdot a^{sw}_{j+\frac{1}{2},i}}{a^{sw}_c}
\end{aligned} \tag{A.4}
$$

## A.3 Reordering of Linear Equations

Define a permutation matrix $R$ of size $n \times n$ with all the values set to 0 except exactly one value of 1 on each row and each column. The reordering of the linear equations of Eq. (2.1) is executed from a mathematical point of view in the following way. Defining $u = R^T v$,

$$
\begin{aligned}
RAu &= Rf, \\
RAR^T v &= Rf.
\end{aligned} \tag{A.5}
$$

It is easy to see that solving (A.5) is mathematically equivalent to solving Eq. (2.1). The computational implementation of the reordering is not executed in practice. In Red-Black and 4-Color relaxation it is done simply by changing the order in which the elements are processed.

## A.4 Minimization of the Frobenius norm for SPAI-0

Using the following definition of the Frobenius norm,

$$\|A\|_F = \sqrt{\sum_{i=1}^{n}\sum_{j=1}^{n} |a_{ij}|^2}, \tag{A.6}$$

the aim is to minimize $g(M)$,

$$g(M) = \|I - MA\|_F = \sqrt{\sum_{k=1}^{n}\left(1 - 2m_k a_{kk} + \sum_{i=1}^{n}(m_k a_{ki})^2\right)}. \tag{A.7}$$

Each $m_k$ value is independent so the minimization of each one of them can be done separately. Additionally, square root is a monotonic function, so in order to find the value of $m_k$ for which the expression reaches a minimum there is no need to take the square root.

In other words, to minimise $g(M)$ we can minimise

$$f(m_k) = 1 - 2m_k a_{kk} + \sum_{i=1}^{n}(m_k a_{ki})^2 = 1 - 2m_k a_{kk} + m_k^2 \sum_{i=1}^{n} a_{ki}^2, \tag{A.8}$$

for $k = 1, ..., n$. The derivative $f'$ of $f$ is

$$f'(m_k) = -2a_{kk} + 2m_k \sum_{i=1}^{n} a_{ki}^2. \tag{A.9}$$

The minimum given by $f'(m_k) = 0$ is reached for

$$m_k = \frac{a_{kk}}{\sum_{i=1}^{n} a_{ki}^2}. \tag{A.10}$$

## A.5 Additional Normalised NN results

Table A.1: Comparison of the convergence rates of SPAI-0 (last line) and the NN trained with different parameters on normalized data. *Pre* and *Post* are the numbers of relaxations performed respectively before and after the coarse grid correction. Columns *NN wins* give the percentage of samples for which NN outperforms SPAI-0. Columns *Conv. rate* show the geometrical average of convergence rates. The structure of the table is the same as the structure of table 4.6

| Method | | | Results of testing | | | |
|---|---|---|---|---|---|---|
| Type | Training param. | | $\nu_1 = 0, \nu_2 = 1$ | | $\nu_1 = 1, \nu_2 = 1$ | |
| | $\nu$ | $\alpha$ | NN wins | Conv. rate | NN wins | Conv. rate |
| NN | 1 | 1 | 3% | 0.7828 | 18.5% | 0.7215 |
| NN | 1 | 2 | 99.5% | 0.7470 | 83.5% | 0.7110 |
| NN | 1 | 4 | 99.5% | 0.7148 | 75% | 0.7068 |
| NN | 1 | 6 | 99.5% | 0.6965 | 64.5% | 0.7068 |
| NN | 1 | 8 | 99.5% | 0.6774 | 53.5 | 0.7071 |
| NN | 1 | 10 | 99.5% | 0.6638 | 45.5% | 0.7088 |
| NN | 2 | 1 | 99% | 0.7591 | 83% | 0.7133 |
| NN | 2 | 2 | 99.5% | 0.7479 | 83.5% | 0.7111 |
| NN | 2 | 4 | 99.5% | 0.7452 | 84% | 0.7107 |
| NN | 2 | 6 | 99.5% | 0.7421 | 84.5% | 0.7104 |
| NN | 2 | 8 | 99.5% | 0.7413 | 85% | 0.7101 |
| NN | 2 | 10 | 99.5% | 0.7403 | 84.5% | 0.7099 |
| NN | 3 | 1 | 99 | 0.7565 | 83% | 0.7125 |
| NN | 3 | 2 | 99.5% | 0.7015 | 70% | 0.7068 |
| NN | 3 | 4 | 99.5% | 0.6420 | 18.5% | 0.7163 |
| NN | 3 | 6 | 96.5% | 0.6842 | 8.5% | 0.7224 |
| NN | 3 | 8 | 75% | 0.7079 | 6.5% | 0.7256 |
| NN | 3 | 10 | 58.5% | 0.7204 | 5.5% | 0.7272 |
| NN | 4 | 1 | 99% | 0.7537 | 84.5% | 0.7121 |
| NN | 4 | 2 | 99.5% | 0.7018 | 70% | 0.7068 |
| NN | 4 | 4 | 99.5% | 0.6836 | 59% | 0.7070 |
| NN | 4 | 6 | 99.5% | 0.6776 | 53.5% | 0.7071 |
| NN | 4 | 8 | 99.5% | 0.6741 | 52% | 0.7076 |
| NN | 4 | 10 | 99.5% | 0.6706 | 52% | 0.7081 |
| SPAI-0 | - | - | - | 0.7753 | - | 0.7189 |

## A.6 Additional results on W cycles

Table A.2: W-cycle for testing on grid size $64\times64$. *Common* refers to 4-Color SOR with identical input parameters. $64 \times 64$ in $NN_{64\times64}$ indicates the size of the grid used for training. $NN_{64\times64} \pm 0.01$ indicates best values after conducting a local search algorithm for training. *Brute force* indicates best values found using a genetic search algorithm for training. The *Improvement* columns indicate the number of testing cycles required to reach the same residual as a proportion of the number of cycles required by two common 4-Color SOR processes (indicated in the columns by 100%). The structure of the table is the same as the structure of table 4.14. The dataset is different.

| Type | 4-Colors values | | | | Convergence Rate | Improvement | |
|---|---|---|---|---|---|---|---|
| Common | 1.00 | 1.00 | 1.00 | 1.00 | 0.3143 | — | 100% |
| Common | ··· | ··· | ··· | ··· | — | — | — |
| Common | 1.080 | 1.080 | 1.080 | 1.080 | 0.2144 | — | 75.15% |
| Common | 1.090 | 1.090 | 1.090 | 1.090 | 0.2097 | 100% | 74.10% |
| Common | 1.100 | 1.100 | 1.100 | 1.100 | 0.2107 | — | 74.31% |
| $NN_{64\times64}$ - exact | 0.756 | 1.119 | 1.119 | 1.052 | 0.1601 | 85.25% | 63.17% |
| $NN_{64\times64} \pm 0.01$ | 0.766 | 1.129 | 1.129 | 1.062 | 0.1525 | 83.05% | 61.54% |
| Brute force | 0.776 | 1.159 | 1.159 | 1.052 | 0.1482 | 81.81% | 60.62% |

Table A.3: Same as Table A.2. The dataset is different.

| Type | 4-Colors values | | | | Convergence Rate | Improvement | |
|---|---|---|---|---|---|---|---|
| Common | 1.00 | 1.00 | 1.00 | 1.00 | 0.3141 | — | 100% |
| Common | ··· | ··· | ··· | ··· | — | — | — |
| Common | 1.080 | 1.080 | 1.080 | 1.080 | 0.2227 | — | 77.09% |
| Common | 1.090 | 1.090 | 1.090 | 1.090 | 0.2187 | 100% | 76.18% |
| Common | 1.100 | 1.100 | 1.100 | 1.100 | 0.2210 | — | 76.70% |
| $NN_{64\times64}$ - exact | 0.756 | 1.119 | 1.119 | 1.052 | 0.1727 | 86.55% | 65.93% |
| $NN_{64\times64} \pm 0.01$ | 0.766 | 1.129 | 1.129 | 1.062 | 0.1626 | 83.68% | 63.75% |
| Brute force | 0.776 | 1.169 | 1.159 | 1.062 | 0.1549 | 81.51% | 62.10% |

Table A.4: Same as Table A.2. The dataset is different.

| Type | 4-Colors values | | | | Convergence Rate | Improvement | |
|---|---|---|---|---|---|---|---|
| Common | 1.00 | 1.00 | 1.00 | 1.00 | 0.2861 | — | 100% |
| Common | ... | ... | ... | ... | — | — | — |
| Common | 1.070 | 1.070 | 1.070 | 1.070 | 0.2006 | — | 77.90% |
| Common | 1.080 | 1.080 | 1.080 | 1.080 | 0.1944 | 100% | 76.39% |
| Common | 1.090 | 1.090 | 1.090 | 1.090 | 0.1993 | — | 77.59% |
| $NN_{64\times64}$ - exact | 0.756 | 1.119 | 1.119 | 1.052 | 0.1458 | 85.07% | 64.99% |
| $NN_{64\times64} \pm 0.01$ | 0.766 | 1.129 | 1.129 | 1.052 | 0.1324 | 81.02% | 61.90% |
| Brute force | 0.766 | 1.129 | 1.129 | 1.052 | 0.1324 | 81.02% | 61.90% |

## A.7 Results on F cycles

Table A.5: F-cycle for testing on grid size $64 \times 64$. *Common* refers to 4-Color SOR with identical input parameters. $64 \times 64$ in $NN_{64\times64}$ indicates the size of the grid used for training. $NN_{64\times64}\pm0.01$ indicates best values after conducting a local search algorithm for training. *Brute force* indicates best values found using a genetic search algorithm for training. The *Improvement* columns indicate the number of testing cycles required to reach the same residual as a proportion of the number of cycles required by two common 4-Color SOR processes (indicated in the columns by 100%). The structure of the table is the same as the structure of table A.2.

| Type | 4-Colors values | | | | Convergence Rate | Improvement | |
|---|---|---|---|---|---|---|---|
| Common | 1.00 | 1.00 | 1.00 | 1.00 | 0.3010 | — | 100% |
| Common | ... | ... | ... | ... | — | — | — |
| Common | 1.070 | 1.070 | 1.070 | 1.070 | 0.2008 | — | 74.79% |
| Common | 1.080 | 1.080 | 1.080 | 1.080 | 0.1953 | 100% | 73.52% |
| Common | 1.090 | 1.090 | 1.090 | 1.090 | 0.1961 | — | 73.71% |
| $NN_{64\times64}$ - exact | 0.756 | 1.119 | 1.119 | 1.052 | 0.1408 | 83.31% | 61.25% |
| $NN_{64\times64} \pm 0.01$ | 0.766 | 1.129 | 1.129 | 1.052 | 0.1343 | 81.34% | 59.80% |
| Brute force | 0.766 | 1.129 | 1.139 | 1.052 | 0.1340 | 81.27% | 59.75% |

Table A.6: Same as Table A.5. The dataset is different.

| Type | 4-Colors values | | | | Convergence Rate | Improvement | |
|---|---|---|---|---|---|---|---|
| Common | 1.00 | 1.00 | 1.00 | 1.00 | 0.3144 | — | 100% |
| Common | ··· | ··· | ··· | ··· | — | — | — |
| Common | 1.080 | 1.080 | 1.080 | 1.080 | 0.2144 | — | 75.15% |
| Common | 1.090 | 1.090 | 1.090 | 1.090 | 0.2098 | 100% | 74.09% |
| Common | 1.100 | 1.100 | 1.100 | 1.100 | 0.2107 | — | 74.30% |
| $NN_{64\times64}$ - exact | 0.756 | 1.119 | 1.119 | 1.052 | 0.1600 | 85.23% | 63.15% |
| $NN_{64\times64} \pm 0.01$ | 0.766 | 1.129 | 1.129 | 1.052 | 0.1525 | 83.05% | 61.53% |
| Brute force | 0.766 | 1.159 | 1.149 | 1.052 | 0.1482 | 81.80% | 60.61% |

## A.8 Results on V cycles

Table A.7: V-cycle for testing on grid size $64 \times 64$. *Common* refers to 4-Color SOR with identical input parameters. $64 \times 64$ in $NN_{64\times64}$ indicates the size of the grid used for training. $NN_{64\times64} \pm 0.01$ indicates best values after conducting a local search algorithm for training. *Brute force* indicates best values found using a genetic search algorithm for training. The *Improvement* columns indicate the number of testing cycles required to reach the same residual as a proportion of the number of cycles required by two common 4-Color SOR processes (indicated in the columns by 100%). The structure of the table is the same as the structure of table A.2.

| Type | 4-Colors values | | | | Convergence Rate | Improvement | |
|---|---|---|---|---|---|---|---|
| Common | 1.00 | 1.00 | 1.00 | 1.00 | 0.3244 | — | 100% |
| Common | ··· | ··· | ··· | ··· | — | — | — |
| Common | 1.100 | 1.100 | 1.100 | 1.100 | 0.2282 | — | 76.20% |
| Common | 1.110 | 1.110 | 1.110 | 1.110 | 0.2280 | 100% | 76.15% |
| Common | 1.120 | 1.120 | 1.120 | 1.120 | 0.2344 | — | 77.60% |
| $NN_{64\times64}$ - exact | 0.7556 | 1.119 | 1.119 | 1.042 | 0.2430 | 104.50% | 79.58% |
| $NN_{64\times64} \pm 0.01$ | 0.766 | 1.129 | 1.129 | 1.062 | 0.2261 | 99.42% | 75.71% |
| Brute force | 0.766 | 1.169 | 1.169 | 1.112 | 0.1839 | 87.31% | 66.49% |